\documentclass{article}
\usepackage{ijcai23}

\usepackage{times}
\usepackage{soul}
\usepackage{url}
\usepackage[hidelinks]{hyperref}
\usepackage[utf8]{inputenc}
\usepackage[small]{caption}
\usepackage{graphicx}
\usepackage{amsmath}
\usepackage{amsthm}
\usepackage{booktabs}
\usepackage{tabularx}
\usepackage{makecell}
\usepackage{algorithm}
\usepackage{algorithmic}
\usepackage[switch]{lineno}
\usepackage{amsfonts}
\usepackage{amssymb}
\usepackage{xspace}
\usepackage{tikz}
\usepackage{pgfplots}
\usepackage{pgfplotstable}
\usepackage{cleveref}
\pgfplotsset{compat=1.16}
\usetikzlibrary{shapes,shapes.geometric,arrows,arrows.meta,fit,calc,positioning,automata,angles,quotes}
\usetikzlibrary{pgfplots.statistics}
\usepgfplotslibrary{fillbetween} 
\usepackage{subcaption}

\definecolor{darkgreen}{rgb}{0,0.6,0}
\usepackage{nicefrac}

\usepackage{todonotes}
\usepackage{xcolor}

\newcommand{\citet}[1]{\citeauthor{#1}~\shortcite{#1}}

\definecolor{red}{rgb}{0.745,0.192,0.102}
\definecolor{myred}{RGB}{158, 37, 16}
\definecolor{mygreen}{RGB}{8, 138,95}
\definecolor{mydarkblue}{RGB}{8,59,158}
\definecolor{myyellow}{RGB}{158, 114,16}
\definecolor{grey}{gray}{0.85}
\definecolor{lightgrey}{gray}{0.95}

\renewcommand{\subsubsection}[1]{\leavevmode\unskip\\\noindent\textbf{#1}}

\newcommand{\cD}{\ensuremath{\mathcal{D}}\xspace}

\newcommand{\cU}{\ensuremath{\mathcal{U}}\xspace}

\newcommand{\Post}{\mathit{Post}}

\makeatletter
\newcommand{\oset}[3][-.2ex]{%
  \mathrel{\mathop{#3}\limits^{
    \vbox to#1{\kern-2\ex@
    \hbox{$\scriptstyle#2$}\vss}}}}
\makeatother

\newcommand{\Nat}{\mathbb{N}}

\newcommand{\Real}{\mathbb{R}}

\newcommand{\norm}[1]{\lVert#1\rVert}

\newcommand{\prob}{\ensuremath{\mathbb{P}}}

\newcommand{\Dist}{\Delta}

\newcommand{\Exp}{\mathbb{E}}

\newtheorem{definition}{Definition}
\newtheorem{theorem}{Theorem}
\newtheorem{lemma}{Lemma}

\newtheorem{corollary}{Corollary}

\newcommand{\eg}{\emph{e.g.}\xspace}
\newcommand{\ie}{\emph{i.e.}\xspace}

\newcommand{\Rmax}{\ensuremath{R_\text{max}}}

\newcommand{\pii}{\ensuremath{\pi_I}}
\newcommand{\pib}{\ensuremath{\pi_b}}
\newcommand{\data}{\ensuremath{\cD}}
\newcommand{\given}{\ensuremath{\,|\,}}
\newcommand{\mlemdp}{{\ensuremath{\tilde{M}}}}
\newcommand{\mleP}{\ensuremath{\tilde{P}}}

\newcommand{\cnt}{\ensuremath{\#}}

\newcommand{\Ind}{\ensuremath{\mathbb{I}}}

\newcommand{\tsMDP}{2sMDP\xspace}
\newcommand{\twosMDP}{two-successor MDP\xspace}
\newcommand{\spibb}{{\text{SPIBB}}}

\newcommand{\Smain}{\ensuremath{S}}
\newcommand{\Saux}{\ensuremath{S_\text{aux}}}

\newcommand{\Mts}{\ensuremath{{M^\ts}}}
\newcommand{\pits}{\ensuremath{{\pi^\ts}}}
\newcommand{\datats}{\ensuremath{{\data^\ts}}}

\newcommand{\mlemdpts}{{\ensuremath{\tilde{M}^\ts}}}
\newcommand{\ts}{\ensuremath{\mathsf{2s}}}

\newcommand{\Nmin}{\ensuremath{N_\wedge}}
\newcommand{\Vmax}{\ensuremath{V_{max}}}

\title{More for Less: Safe Policy Improvement \\With Stronger Performance Guarantees}

\author{
Patrick Wienh\"{o}ft$^{1,2}$\footnote{Shared First Authorship.}\and
Marnix Suilen$^3$\footnotemark[1]\and
Thiago D. Sim\~{a}o$^3$\and\\
Clemens Dubslaff$^{4,2}$\and
Christel Baier$^{1,2}$\And
Nils Jansen$^3$
\affiliations
$^1$Department of Computer Science, Technische Universit\"at Dresden, Dresden, Germany\\
$^2$Centre for Tactile Internet with Human-in-the-Loop (CeTI) \\
$^3$Department of Software Science, Radboud University, Nijmegen, The Netherlands\\
$^4$Eindhoven University of Technology, Eindhoven, The Netherlands
 \emails
 \{patrick.wienhoeft, christel.baier\}@tu-dresden.de \\
 \{m.suilen, t.simao, n.jansen\}@science.ru.nl \\
 c.dubslaff@tue.nl
}

\begin{document}

\maketitle

\begin{abstract}

In an offline {reinforcement learning} setting, the {safe policy improvement} (SPI) problem aims to improve the performance of a behavior policy according to which sample data has been generated.
State-of-the-art approaches to SPI require a high number of samples to provide practical probabilistic guarantees on the improved policy's performance.
We present a novel approach to the SPI problem that provides the means to require less data for such guarantees. 
Specifically, to prove the correctness of these guarantees, we devise implicit transformations on the data set and the underlying environment model that serve as theoretical foundations to derive tighter improvement bounds for SPI.
Our empirical evaluation, using the well-established {SPI with baseline bootstrapping} (SPIBB) algorithm, on standard benchmarks shows that our method indeed significantly reduces the sample complexity of the SPIBB algorithm.

\end{abstract}
\section{Introduction}\label{sec:introduction}

\emph{Markov decision processes} (MDPs) are the standard model for sequential decision-making 
under uncertainty~\cite{DBLP:books/wi/Puterman94}. 
\emph{Reinforcement learning} (RL) solves such decision-making problems, in particular when the environment dynamics are unknown~\cite{DBLP:books/lib/SuttonB98}.

In an \emph{online} RL setting, an agent aims to learn a decision-making policy that maximizes the expected accumulated 
reward by interacting with the environment and observing feedback, typically in the form of information about the environment state and reward. 
While online RL has shown great performance in solving hard problems~\cite{DBLP:journals/nature/MnihKSRVBGRFOPB15,Silver2018generalRL}, 
the assumption that the agent can always directly interact with the environment is not always realistic.
In real-world applications such as robotics or healthcare, direct interaction can be impractical or dangerous~\cite{DBLP:journals/corr/abs-2005-01643}.
Furthermore, alternatives such as simulators or digital twins may not be available or insufficiently 
capture the nuances of the real-world application for reliable learning~\cite{DBLP:journals/jair/RamakrishnanKDH20,DBLP:conf/ssci/ZhaoQW20}. 

\emph{Offline RL} (or batch RL)~\cite{DBLP:books/sp/12/LangeGR12} mitigates this concern by restricting the agent to have only 
access to a fixed data set of past interactions.
As a common assumption, the data set has been generated by a so-called \emph{behavior policy}.
An offline RL algorithm  aims to produce a new policy without further interactions with the environment~\cite{DBLP:journals/corr/abs-2005-01643}.
Methods that can reliably improve the performance of a policy are key in 
(offline) RL.

\emph{Safe policy improvement} (SPI) algorithms address this challenge by providing (probabilistic) correctness guarantees on the reliable improvement of policies~\cite{DBLP:conf/icml/ThomasTG15,Petrik2016}.
These guarantees depend on the size of the data set and usually adhere to a conservative bound on the minimal amount of samples required.
Since this bound often turns out to be too large for practical applications of SPI,
it is instead turned into a hyperparameter~(see, \eg,~\cite{DBLP:conf/icml/LarocheTC19}).
The offline nature of SPI prevents further data collection, which steers the key requirements of SPI in practical settings: (1) exploit the data set as efficiently as possible and (2) compute better policies from smaller data sets. 

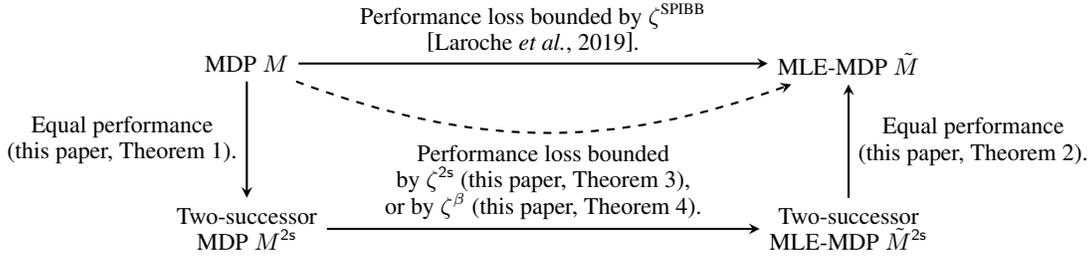
\begin{figure*}[t]
    \centering
    \begin{tikzpicture}[thick,scale=1, every node/.style={font=\small}]

\node (M) at (0,0) {MDP $M$};
\node[align=center] (Mp) at (0,-2.2) {Two-successor \\ MDP $\Mts$};
\node (Mhat) at (8,0) {MLE-MDP $\mlemdp$};
\node[align=center] (Mphat) at (8,-2.2) {Two-successor \\ MLE-MDP $\mlemdpts$};

\draw [-stealth] (M) -- node [align=center, midway, left] {Equal performance\\ (this paper, \Cref{thm:mdp_transformation}).} (Mp);
\draw [-stealth] (M) -- node [align=center, midway, above] {Performance loss bounded by $\zeta^\text{SPIBB}$\\ \cite{DBLP:conf/icml/LarocheTC19}.} (Mhat);
\draw [-stealth] (Mp) -- node [align=center, midway, above] {Performance loss bounded\\ by $\zeta^\ts$ (this paper, Theorem~\ref{thm:2smdp_spi}),\\ or by $\zeta^{\beta}$ (this paper, Theorem~\ref{thm:beta-zeta-bound}).} (Mphat);
\draw[-stealth] (M) edge[dashed, bend right=20] (Mhat);

\draw [-stealth] (Mphat) -- node [align=center, midway, right] {Equal performance\\ (this paper, \Cref{thm:data_transformation}).} (Mhat);

\end{tikzpicture}
    \caption{Overview of our approach. The solid arrows indicate how the full derivation of the improvement guarantees is done, while the dashed line indicates that the transformations are only used in the proofs and that in practice, we can immediately use $\zeta^\ts$ or $\zeta^\beta$ as bounds.}
    \label{fig:approach-overview}
\end{figure*}

\subsubsection{Contributions.}
Our contribution provides the theoretical foundations to improve the understanding of SPI algorithms in general.
Specifically, in a general SPI setting, we can guarantee a higher performance for significantly less data. 
Equivalently, we can allow the same amount of data and consequently provide significantly less performance guarantees.
Our main technical contribution is an transformation of the underlying MDP model into a \emph{\twosMDP} (\tsMDP) along with adjustments to the data set, that allows us to prove these tighter bounds.
A \tsMDP is an MDP where each state-action pair has at most two successors,
hence limiting the branching factor of an MDP to only two.
These transformations preserve (the optimal) performance of policies and are reversible.
In the context of SPI these transformations are implicit, \ie, do not have to be computed explicitly.
Hence, we are able to apply standard SPI algorithms such as SPI with baseline bootstrapping (SPIBB)~\cite{DBLP:conf/icml/LarocheTC19}, and use our novel improvement guarantees without any algorithmic changes necessary, as also illustrated in Figure~\ref{fig:approach-overview}.

Following the theoretical foundations for the MDP and data set transformations~(\Cref{sec:transformations}), 
we present two different methods to compute the new performance guarantees (\Cref{sec:SPI_tsMDP}). 
The first uses Weissman's bound~\cite{Weissman}, as also used in, \eg, standard SPIBB, 
while the second uses the inverse incomplete beta function~\cite{Temme92}.
Our experimental results show a significant reduction in the amount of data 
required for equal performance guarantees~(\Cref{sec:evaluation}).
Concretely, where the number of samples required at each state-action pair of standard 
SPIBB grows linearly in the number of states, our approach only grows logarithmic in the number of states for both methods.
We also demonstrate the impact on three well-known benchmarks in practice by comparing 
them with standard SPIBB across multiple hyperparameters.

\section{Preliminaries}\label{sec:preliminaries}
Let $X$ be a finite set. We denote the number of elements in $X$ by $|X|$.
A discrete probability distribution over $X$ is a function 
$\mu \colon X \to [0,1]$ where $\sum_{x \in X} \mu(x) = 1$.
The set of all such distributions is denoted by $\Dist(X)$.
The $L_1$-distance between two probability distributions $\mu$ and $\sigma$ is defined as 
$
\lVert \mu - \sigma \rVert_1 = \sum_{x \in X} |\mu(x) - \sigma(x)|
$.
We write $[m:n]$ for the set of natural numbers $\{m, \dots, n\} \subset \Nat$, 
and $\Ind[x{=}x']$ for the indicator function, returning $1$ if 
$x = x'$ and $0$ otherwise.

\begin{definition}[MDP]
A \emph{Markov decision process} (MDP) is a tuple $M=(S,A,\iota,P,R,\gamma)$, where $S$ 
and $A$ are finite sets of states and actions, respectively, $\iota\in S$ an initial state, $P \colon S \times A \rightharpoonup \Dist(S)$ is the (partial) transition function, $R \colon S \times A \rightharpoonup [-\Rmax, \Rmax]$ is the reward function bounded by some known value $\Rmax \in \Real$, and $\gamma \in (0,1) \subset \Real$ is the discount factor.
\end{definition}

We say that an action $a$ is \emph{enabled} in state $s$ if $P(s,a)$ is defined.
We write $P(s' \given s,a)$ for the transition probability $P(s,a)(s')$, and $\Post_M(s,a)$ for the set of successor states reachable with positive probability from the state-action pair $(s,a)$ in $M$.
A \emph{path} in $M$ is a finite sequence $\langle s_1, a_1, \ldots, a_{n-1}, s_n \rangle \in (S \times A)^* \times S$ where 
$s_i\in\Post_M(s_{i-1},a_{i-1})$ for all $i\in[2{:}n]$.
The probability of following a path $\langle s_1, a_1, \dots, a_{n-1}, s_n \rangle$ in the MDP~$M$ given a deterministic sequence of actions is written as $\prob_M(\langle s_1, a_1, \dots, a_{n-1}, s_n \rangle)$ and can be computed by repeatedly applying the transition probability function, \ie, $\prob_M(\langle s_1, a_1, \dots, a_{n-1}, s_n \rangle)=\prod_{i=1}^{{n-1}}P(s_{i+1}\given s_i,a_i)$.

A memoryless stochastic \emph{policy} for $M$ is a function $\pi \colon S \to \Dist(A)$.
The set of such policies is $\Pi$.
The goal is to find a policy maximizing the expected discounted reward
\[
\max_{\pi \in \Pi} \Exp \left[ \sum_{t=1}^{\infty} \gamma^t r_t \right],
\]
where $r_t$ is the reward the agent collects at time $t$ when following policy $\pi$ in the MDP.

We write $V_M^\pi(s)$ for the state-based \emph{value function} of an MDP $M$ under a policy $\pi$.
Whenever clear from context, we omit $M$ and $\pi$.
The value of a state $s$ in an MDP $M$ is the least solution of the Bellman equation and can be computed by, \eg, value iteration~\cite{DBLP:books/wi/Puterman94}.
The \emph{performance} $\rho(\pi, M)$ of a policy $\pi$ in an MDP $M$ is defined as the value in the initial state $\iota \in S$, \ie, $\rho(\pi, M) = V_M^\pi(\iota)$.

\section{Safe Policy Improvement}\label{sec:SPI}
In \emph{safe policy improvement} (SPI), we are given an MDP $M$ with an unknown transition function, a policy $\pib$, also known as 
the \emph{behavior policy}, and a data set $\data$ of paths in $M$ under $\pib$.
The goal is to derive a policy $\pii$ from $\pib$ and $\data$ that with high probability $1{-}\delta$
guarantees to \textit{improve} $\pib$ on $M$ up to an \emph{admissible performance loss} $\zeta$.
That is, the performance of $\pii$ is at least that of $\pib$ tolerating an error of $\zeta$:
\begin{equation}\label{eq:spi:performance-improvement}
   \rho(\pii, M) \geq \rho(\pib, M) - \zeta. 
\end{equation}

\subsection{Maximum Likelihood Estimation}
We use \emph{maximum likelihood estimation} (MLE) to derive an MLE-MDP $\tilde{M}$ from the data set $\data$.
For a path $\rho \in \data$, let $\cnt_\rho(s,a)$ and $\cnt_\rho(s,a,s')$ be the number of (sequential) occurrences of a state-action pair $(s,a)$ and a transition $(s,a,s')$ in $\rho$, respectively. We lift this notation the level of the data set $\data$ by defining $\cnt_\data(s,a)=\sum_{\rho\in\data} \cnt_\rho(s,a)$ and $\cnt_\data(s,a,s')=\sum_{\rho\in\data} \cnt_\rho(s,a,s')$.
\begin{definition}[MLE-MDP]
The \emph{maximum likelihood MDP} (MLE-MDP) of an MDP $M = (S,A,\iota,P,R,\gamma)$ and data set $\data$ is a tuple $\mlemdp = (S,A,\iota,\mleP, R, \gamma)$ where $S,\iota,A,R,$ and $\gamma$ are as in $M$ and the transition function is estimated from $\data$:
\begin{align*}
    \mleP(s' \given s,a) = \frac{\cnt_\data(s,a,s')}{\cnt_\data(s,a)}.
\end{align*}
\end{definition}

Let $e \colon S {\times} A \to \Real$ be an error function.
We define $\Xi_e^{\mlemdp}$ as the set of MDPs $M'$ 
that are close to $\mlemdp$, \ie, %
where for all state-action pairs $(s,a)$ the $L_1$-distance 
between the transition function $P'(\cdot \given s,a)$ and $\mleP(\cdot \given s,a)$ is at most $e(s,a)$:
\[
\Xi_e^{\mlemdp} = \{M' \mid \forall (s,a). \lVert P'(\cdot \given s,a) - \mleP(\cdot \given s, a) \rVert_1 \leq e(s,a) \}.
\]
SPI methods aim at defining the error function $e$ in such a way that $\Xi_{e}^{\mlemdp}$ contains the true MDP $M$ with high probability $1 - \delta$.
Computing a policy that is an improvement over the behavior policy for all MDPs in this set 
then also guarantees an improved policy for the MDP $M$ with high probability $1-\delta$~\cite{Petrik2016}.
The amount of data required to achieve a $\zeta^{\text{SPI}}$-approximately safe policy improvement with probability $1-\delta$ (recall \Cref{eq:spi:performance-improvement}) for all state-action pairs has been established by~\citeauthor{DBLP:conf/icml/LarocheTC19}~(\shortcite{DBLP:conf/icml/LarocheTC19}) as
\begin{equation}\label{eq:spi-data-bound}
    \cnt_\data(s,a) \geq \Nmin^{\text{SPI}} = \frac{8 \Vmax^2}{({\zeta^{\text{SPI}}})^2 (1-\gamma)^2} \log \frac{2 |S||A|2^{|S|}}{\delta}.
\end{equation}
Intuitively, if the data set $\data$ satisfies the constraint in Equation~\ref{eq:spi-data-bound}, the MLE-MDP estimated from $\data$ will be close enough to the unknown MDP $M$ used to obtain $\data$. 
To this end, it would be likely that a policy in the MLE-MDP with better performance will also have a better performance in $M$.

\subsection{SPI with Baseline Bootstrapping}\label{subsec:spibb}
The constraint in \Cref{eq:spi-data-bound} has to hold for all state-action pairs in order to guarantee a $\zeta$-approximate improvement and thus requires a large data set with good coverage of the entire model.
SPI with baseline bootstrapping (SPIBB)~\cite{DBLP:conf/icml/LarocheTC19} relaxes this requirement by only changing the behavior policy in those pairs for which the data set contains enough samples and follows the behavior policy otherwise.
Specifically, state-action pairs with less than  $\Nmin^\text{SPIBB}$ samples
are collected in a set of \emph{unknown} state-action pairs $\cU$:
\[
\cU = \{ (s,a) \in S \times A \mid \cnt_\data(s,a) \leq \Nmin^\text{SPIBB} \}.
\]
SPIBB then determines an improved policy $\pii$ similar to (standard) SPI, except that if $(s,a) \in \cU$, $\pii$ is required to follow the behavior policy $\pib$:
\[
\forall (s,a) \in \cU. \, \pii(a \given s) = \pib(a \given s).
\]
Then, $\pii$ is an improved policy as in \Cref{eq:spi:performance-improvement}, where $\Nmin^\spibb$ is treated as a hyperparameter and $\zeta$ is given by
\begin{align*}
\zeta^\spibb &= \frac{4\Vmax}{1-\gamma} \sqrt{\frac{2}{\Nmin^\spibb} \log \frac{2|S||A|2^{|S|}}{\delta}} \\
& \qquad \qquad \qquad \qquad - \rho(\pii, \mlemdp) + \rho(\pib, \mlemdp).
\end{align*}

We can rearrange this equation to compute the number of necessary samples for a $\zeta^\spibb$-approximate improvement. 
As $\rho(\pii,\mlemdp)$ is only known at runtime, we have to employ an under-approximation
$\rho(\pi_b,\mlemdp)$ to a priori compute
\[
 \Nmin^\spibb = \frac{32 \Vmax^2}{{(\zeta^\spibb})^2 (1-\gamma)^2} \log \frac{2 |S||A|2^{|S|}}{\delta}.
\]

Thus, the sample size constraint $\Nmin^\text{SPIBB}$ grows approximately linearly in terms of the size of the MDP.
The exponent of the term $2^{|S|}$ is an over-approximation of the maximum branching factor of the MDP, since worst-case, the MDP can be fully connected.
In the following Section, we present our approach to limit the branching factor of an MDP.
After that, we present two methods that exploit this limited branching factor to derive improved sampling size constraints for SPI that satisfy the same guarantees.

\section{Tighter Improvement Bounds for SPI}\label{sec:transformations}
In the following, we present the technical construction of \twosMDP{}s and the data set transformation that allows us to derive the tighter performance guarantees in SPI.

\subsection{From MDP to Two-Successor MDP}\label{subsec:mdp_transformation}
A \emph{\twosMDP} (\tsMDP) is an MDP $\Mts$ where each state-action pair $(s,a)$ has at most two possible successors states, \ie, 
$|\Post_{M^{2s}}(s,a)| \leq 2$.
To transform an MDP $M = (S,A,\iota,P,R,\gamma)$ into a \tsMDP, we introduce a set 
of \emph{auxiliary} states $\Saux$ along with the \emph{main} states $\Smain$ of the MDP $M$. 
Further, we include an additional action $\tau$ and adapt the probability and reward functions towards a \tsMDP $\Mts = (\Smain \cup \Saux, A \cup \{\tau\}, \iota, P^\ts, R^\ts, \gamma^\ts)$.

For readability, we now detail the transformation for a fixed state-action pair $(s,a)$ with three or more successors.
The transformation of the whole MDP follows from repeatedly applying this transformation to all such state-action pairs.

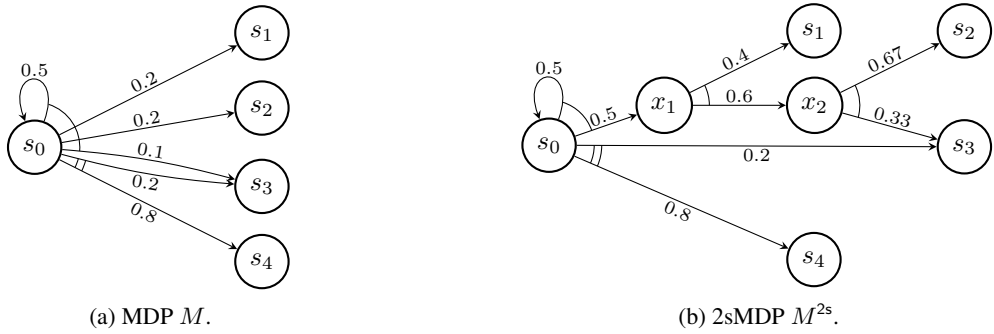
\begin{figure*}
     \centering
     \begin{subfigure}[b]{0.4\textwidth}
         \centering
         \begin{tikzpicture}
\tikzstyle{round}=[thick,draw=black,circle]

    \node[round] (s0) {$s_0$};
    \node[round,above right=10mm and 25mm of s0] (s1) {$s_1$};
    \node[round,above right=0mm and 25mm of s0] (s2) {$s_2$};
    \node[round, below right=0mm and 25mm of s0] (s3) {$s_3$};
    \node[round,below right=10mm and 25mm of s0] (s4) {$s_4$};

    \draw[-stealth] (s0) -- node[above=-0.8mm,sloped] {\scriptsize $0.2$} (s1);
    \draw[-stealth] (s0) -- node[above=-0.8mm,sloped] {\scriptsize $0.2$} (s2);
    \draw[-stealth] (s0) to [bend left=5] coordinate[pos=0.1](bb) node[above=-0.8mm,sloped] {\scriptsize $0.1$}  (s3);
    \draw[-stealth] (s0) [out=70,in=110,loop] to coordinate[pos=0.1](aa) node[above=-0.8mm,sloped] {\scriptsize $0.5$}  (s0);
    \path pic[draw, angle radius=6mm,angle eccentricity=1.2] {angle = bb--s0--aa};
    
    \draw[-stealth] (s0) -- node[below=-0.8mm,sloped] {\scriptsize $0.8$}  (s4);
    \draw[-stealth] (s0) to [bend right=5] coordinate[pos=0.1](cc) node[below=-0.8mm,sloped] {\scriptsize $0.2$}  (s3);
    \path pic[draw, angle radius=6mm,angle eccentricity=1.2] {angle = s4--s0--cc};
    \path pic[draw, angle radius=7mm,angle eccentricity=1.2] {angle = s4--s0--cc};

\end{tikzpicture}
         \caption{MDP $M$.}
         \label{fig:mdp}
     \end{subfigure}
     \begin{subfigure}[b]{0.5\textwidth}
         \centering
         \begin{tikzpicture}
\tikzstyle{round}=[thick,draw=black,circle]

    \node[round] (s0) {$s_0$};
    \node[round,above right=10mm and 30mm of s0] (s1) {$s_1$};
    \node[round,above right=0mm and 10mm of s0] (x1) {$x_1$};
    \node[round,above right=10mm and 50mm of s0] (s2) {$s_2$};
    \node[round,above right=0mm and 30mm of s0] (x2) {$x_2$};
    \node[round, below right=-5mm and 50mm of s0] (s3) {$s_3$};
    \node[round,below right=10mm and 30mm of s0] (s4) {$s_4$};

    \draw[-stealth] (s0) -- node[above=-0.8mm,sloped] {\scriptsize $0.5$} (x1);
    \draw[-stealth] (x1) -- node[above=-0.8mm,sloped] {\scriptsize $0.4$} (s1);
    \draw[-stealth] (x1) -- node[above=-0.8mm,sloped] {\scriptsize $0.6$} (x2);
    \draw[-stealth] (s0) [out=70,in=110,loop] to coordinate[pos=0.1](aa) node[above=-0.8mm,sloped] {\scriptsize $0.5$}  (s0);
     \draw[-stealth] (x2) -- node[above=-0.8mm,sloped] {\scriptsize $0.67$} (s2);
      \draw[-stealth] (x2) -- node[above=-0.8mm,sloped] {\scriptsize $0.33$} (s3);
    \path pic[draw, angle radius=6mm,angle eccentricity=1.2] {angle = x1--s0--aa};
	\path pic[draw, angle radius=6mm,angle eccentricity=1.2] {angle = x2--x1--s1};
	\path pic[draw, angle radius=6mm,angle eccentricity=1.2] {angle = s3--x2--s2};
    
    \draw[-stealth] (s0) -- node[below=-0.8mm,sloped] {\scriptsize $0.8$}  (s4);
    \draw[-stealth] (s0) --node[below=-0.8mm,sloped] {\scriptsize $0.2$}  (s3);
    \path pic[draw, angle radius=6mm,angle eccentricity=1.2] {angle = s4--s0--s3};
    \path pic[draw, angle radius=7mm,angle eccentricity=1.2] {angle = s4--s0--s3};

\end{tikzpicture}
         \caption{\tsMDP $\Mts$.}
         \label{fig:2smdp}
     \end{subfigure}
        \caption{Example for a transformation from an MDP to a \tsMDP, where the single and double arc indicate different actions.}
        \label{fig:mdp_transformation}
\end{figure*}

We enumerate the successor states of $(s,a)$, \ie, $\Post_M(s,a) = \{ s_1, \dots, s_k \}$ and define $p_i = P(s_i \given s,a)$ for all $i=1,\dots,k$.
Further, we introduce $k-2$ auxiliary states $\Saux^{s,a} = \{x_2, \dots, x_{k-1}\}$, each
with one available action with a binary outcome. Concretely, the two possible outcomes in state $x_i$ are ``move to state $s_i$'' or ``move to one of the states $s_{i+1},\dots,s_k$'' where the latter is represented by moving to an auxiliary state $x_{i+1}$,
unless $i=k-1$ in which case we immediately move to $s_k$.
Formally, the new transition function $P^\ts(\cdot \given s,a)$ is:
\begin{align*}
    P^\ts(s_1 \given s,a) &= p_1, \quad P^\ts(x_2 \given s,a) = 1-p_1.
\end{align*}
For the transition function $P^\ts$ in the auxiliary states we define a new 
action $\tau$ that will be the only enabled action in these states.
For $i > 1$, the transition function $P^\ts$ is then 
\begin{align*}
P^\ts(s_i \given x_i, \tau) & = \frac{p_i}{1-(p_1+\dots +p_{i-1})},\\
P^\ts(x_{i+1} \given x_i, \tau, i < k-1) & = 1-\frac{p_i}{1-(p_1+\dots +p_{i-1})},\\
P^\ts(s_k \given x_{k-1}, \tau) & = 1-\frac{p_k}{1-(p_{i-1}+p_k)}.
\end{align*}
An example of this transformation is shown in Figure~\ref{fig:mdp_transformation}, where Figure~\ref{fig:mdp} shows the original MDP and Figure~\ref{fig:2smdp} shows the resulting \tsMDP.
As we introduce $|\Post(s,a)|$ auxiliary states for a state-action pair $(s,a)$, and $k\leq|\Smain|$ in the worst-case of a fully connected MDP, we can bound the number of states in the 2sMDP by $|\Smain \cup \Saux|\leq |\Smain| + |\Smain||A|(|\Smain|-2) \leq |\Smain|^2|A|$. 
Note that we did not specify a particular order for the enumeration of the successor states. 
Further, other transformations utilizing auxiliary states with a different structure (\eg, a balanced binary tree) are possible. 
However, neither the structure of the auxiliary states, nor the order of successor states changes the total number of states in the 2sMDP, which is the deciding factor for the application of this transformation in the context of SPI algorithms.

The extension of the reward function is straightforward, i.e., the agent receives 
the same reward as in the original MDP when in main states and no reward when in auxiliary states:
\[
R^\ts(s,a) = \begin{cases}
    R(s,a) & s \in \Smain, a \in A,\\
    0 & \text{otherwise}.
\end{cases}
\]

Any policy $\pi$ for the MDP $M$ can be extended into a policy $\pits$ for the \tsMDP $\Mts$ by copying $\pi$ for states in $\Smain$ and choosing $\tau$ otherwise:
\[
\pits(a \given s) = \begin{cases}
    \pi(a \given s) & s \in \Smain, \\
    \Ind[a = \tau] & s \in \Saux.
\end{cases}
\]

Finally, in order to preserve discounting correctly, we introduce a state-dependent discount factor $\gamma^\ts$, such that discounting only occurs in the main states, \ie,
\[
	\gamma^\ts(s) \ = \ \begin{cases}
			\gamma & s\in \Smain,\\
			1	& s\in \Saux.
	\end{cases}
\]
This yields the following value function for the \tsMDP $\Mts$:
\begin{align*}
    V_\Mts^\pits(s)  &= 
		\sum\limits_{a \in A} \pits(a \given s) \Big( R^\ts(s,a) \\
  &\qquad + \gamma^\ts(s) \sum\limits_{s' \in S} P^\ts(s' \given s, a) V_\Mts^\pits(s') \Big).
\end{align*}
The performance of policy $\pits$ on $\Mts$ uses the value function defined above 
and is denoted by $\rho^\ts(\pits, \Mts) = V_\Mts^\pits(\iota)$, for the initial state $\iota \in S$. 
Our transformation described above, together with the adjusted value function, indeed preserves the 
performance of the original MDP and policy:

\begin{theorem}[Preservation of transition probabilities]\label{thm:mdp_transformation}
    For every transition $(s,a,s')$ in the original MDP $M$, there exists a unique path $\langle s, a, x_2, \tau, \dots,  x_i, \tau, s' \rangle$ in the \tsMDP $\Mts$ with the same probability.
    That is,
    \[
    \prob_M(\langle s, a, s' \rangle) = \prob_{\Mts}(\langle s, a, x_2, \tau, \dots, x_i, \tau, s' \rangle).
    \]
\end{theorem}
\noindent \Cref{ap:proofs} provides the proofs of all theoretical results.
\begin{corollary}[Preservation of performance]\label{col:performance-preservation}
    Let $M$ be an MDP, $\pi$ a policy for $M$, and $\Mts$ the \twosMDP with policy $\pits$ constructed from $M$ and $\pi$ as described above.
    Then $\rho(\pi, M) = \rho^\ts(\pits, \Mts)$.
\end{corollary}

\subsection{Data-set Transformation}\label{subsec:dataset-transformation}
In the previous section, we discussed how to transform an MDP into a \tsMDP.
However, for SPI we do not have access to the underlying MDP, but only to a data set $\data$ and the behavior policy $\pib$ used to collect this data.
In this section, we present a transformation similar to the one from MDP to \tsMDP, but now for the data set $\data$.
This data set transformation allows us to estimate a \tsMDP from the transformed data via maximum likelihood estimation (MLE).

We again assume a data set $\data$ of observed states and actions of the form $\data = \langle s_t, a_t \rangle_{t \in [1:m]}$ from an MDP $M$.
We transform the data set $\data$ into a data set $\datats$ that we use to define a two-successor MLE-MDP $\mlemdpts$.
Each sample $(s_t, a_t, s_{t+1})$ in $\data$ is transformed into a set of samples, each corresponding to a path from $s_t$ to $s_{t+1}$ via states in $\Saux$ in $\Mts$.
Importantly, the data set transformation only relies on $\data$ and not on any additional knowledge about $M$.

Similar to the notation in \Cref{sec:SPI}, let $\cnt_{\data}(x)$ denote the number of times $x$ occurs in $\data$. 
For each state-action pair $(s,a) \in \Smain \times A$ we denote its successor states in 
$\mlemdp$ as $\Post_{\mlemdp}(s,a) = \{s_i \given \cnt_\data(s,a,s_i)>0\}$, which are again enumerated by $\{s_1,\dots,s_k\}$.
Similarly as for the MDP transformation, we define $\Post_{\mlemdpts}(s,a) = \Post_{\mlemdp}(s,a)$ 
if $k\leq 2$ and $\Post_{\mlemdpts}(s,a) = \{s_1,x_2\}$ otherwise. 
For auxiliary states $x_i\in\Saux^{s,a}$, we define $\Post_{\mlemdpts}(x_i,\tau) = \{s_i,x_{i+1}\}$ 
for $i<k{-}1$ and $\Post_{\mlemdpts}(x_{k-1},\tau) = \{s_{k-1},s_k\}$.
We then define the transformed data set $\datats$ from $\data$ for each $s \in S$ and $s' \in\Post_{\mlemdpts}(s,a)$ as follows:
\begin{align*}
	&\cnt_{\datats}(s,a,s') = 
	\begin{cases}
		\cnt_\data(s,a,s') & s' \in \Smain, \\
		\sum\limits_{j=2}^{k} \cnt_\data(s,a,s_j) & s'=x_2 \in \Saux^{s,a},  \\
		0 & \text{otherwise.}
	\end{cases}
\end{align*} 
Further, for each $x_i \in \Saux^{s,a}$ and $s' \in\Post_{\mlemdpts}(s,a)$\begin{align*}
	&\cnt_{\datats}(x_i,\tau,s') = 
	\begin{cases}
		\cnt_\data(s,a,s') & s' \in \Smain, \\
		\sum\limits_{j=i+1}^{k} \cnt_\data(s,a,s_j) & s'=x_{i+1} \in \Saux^{s,a},  \\
		0 & \text{otherwise.}
	\end{cases}
\end{align*}
The following preservation results for data generated MLE-MDPs are in the line of \Cref{thm:mdp_transformation}
and \Cref{col:performance-preservation}.
See \Cref{fig:approach-overview} for an overview of the relationships between theorems.
\begin{theorem}[Preservation of estimated transition probabilities]\label{thm:data_transformation}
    Let $\data$ be a data set and $\datats$ be the data set
    obtained by the transformation above. 
    Further, let $\mlemdp$ and $\mlemdpts$
    be the MLE-MDPs constructed from $\data$ and $\datats$, respectively.
    Then for every transition $(s,a,s')$ in $\mlemdp$ there is a unique path $\langle s, a, x_2, \tau, \dots, x_i, \tau, s' \rangle$ in $\mlemdpts$ with the same probability:
    \[
    \prob_{\mlemdp}(\langle s,a, s' \rangle) = \prob_{\mlemdpts}(\langle s, a, x_2, \tau, \dots, x_i, \tau, s' \rangle).
    \]
\end{theorem}
\begin{corollary}[Preservation of estimated performance]\label{col:performance-preservation-mle}
    Let $\mlemdp$ and $\mlemdpts$ be the MLE-MDPs as above, constructed from $\data$ and $\datats$, respectively.
    Further, let $\tilde\pi$ be an arbitrary policy on $\mlemdp$ and $\tilde\pi^\ts$
    the policy that extends $\pi$ for $\mlemdpts$ by choosing $\tau$ in all auxiliary states.
    Then $\rho(\tilde\pi, \mlemdp) = \rho^\ts(\tilde\pi^\ts, \mlemdpts)$.
\end{corollary}

We want to emphasize that while $\cD^\ts$ may contain more samples than $\cD$, it does not yield any additional
information. Rather, instead of viewing each transitions sample as an atomic data point, in $\cD^\ts$ 
transition samples are considered like a multi-step process. E.g, The sample $(s,a,s_3)\in\cD$ would
be transformed into the samples $\{(s,a,x_2),(x_2,\tau,x_3),(x_3,\tau,s_3)\}\in\cD^\ts$ which in the construction of the
MLE-MDP are used to estimate the probabilities  
$P(s'\neq s_1 \given s,a), P(s'\neq s_2 \given s,a,s'\neq s_1)$ and 
$P(s'=s_3 \given s,a,s'\neq s_1,s'\neq s_2)$, respectively. The probabilities of these events are
mutually independent, but when multiplied give exactly $P(s_3\given s,a)$.

\section{SPI in Two-Successor MDPs}\label{sec:SPI_tsMDP}
In this section, we discuss how SPI can benefit from \twosMDP{}s as
constructed following our new transformation presented in \Cref{sec:transformations}.
The dominating term in the bound $N$ obtained by~\cite{DBLP:conf/icml/LarocheTC19} is the branching factor of the MDP, which, without any prior information, has to necessarily be over-approximated by $|S|$ (cf. \Cref{subsec:spibb}).
We use our transformation above to bound the branching factor to $k=2$, which allows us to provide stronger guarantees with the same data set (or conversely, require less data to guarantee a set maximum performance loss).
Note that bounding the branching factor by any other constant can be achieved by a similar transformation as in \Cref{sec:transformations}, but $k=2$ leads to an optimal bound (cf. \Cref{par:k_suc_mdp}). 

Let $\mlemdp$ and $\mlemdpts$ be the MLE-MDPs inferred from data sets $\data$ and $\datats$, respectively.
Further, let $\pi_{\odot}$ and $\pi^{\ts}_{\odot}$ denote the optimal policies in these MLE-MDPs, constrained to the set of policies that follow $\pi_b$ for state-action pairs $(s, a) \in \cU$.
Note that these optimal policies can easily be computed using, \eg, standard value iteration.
First, we show how to improve the admissible performance loss $\zeta$ in SPI on \twosMDP{}s.

\begin{lemma}\label{lemma:2smdp_spi}
Let $\Mts$ be a \twosMDP with behavior policy $\pib$.
Then, $\pi^{\ts}_{\odot}$ is a $\zeta$-approximately safe policy
improvement over $\pi_b$ with high probability $1-\delta$, where:
\begin{align*}
 \zeta &= \frac{4V_{max}}{1-\gamma}\sqrt{\frac{2}{\Nmin} \log \frac{8 \lvert S \rvert \lvert A \rvert}{\delta}} + \tilde{\rho}^\ts, 
\end{align*}
with $\tilde{\rho}^\ts = -\rho^\ts(\pi^{\ts}_{\odot},\mlemdpts)+\rho^\ts(\pi_b,\mlemdpts)$.
\end{lemma}
For a general MDP $M$, we can utilize this result by first applying the transformation from \Cref{subsec:mdp_transformation}.

\begin{theorem}[Weissman-based tighter improvement guarantee]\label{thm:2smdp_spi}
Let $M$ be an MDP with behavior policy $\pi_b$.
Then, $\pi_{\odot}$ is a $\zeta^\ts$-approximate safe policy
improvement over $\pi_b$ with high probability $1-\delta$, where:
\[ 
\zeta^\ts = \frac{4V_{max}}{1-\gamma}\sqrt{\frac{2}{\Nmin^\ts} \log \frac{8 \lvert S \rvert^2 \lvert A \rvert^2}{\delta}} -\rho(\pi_{\odot},\mlemdp)+\rho(\pi_b,\mlemdp). 
\]
\end{theorem}
As for $\zeta^\spibb$, we can rearrange the equation to compute the number of necessary samples for a $\zeta^\ts$-safe improvement:
\begin{align*}
    \Nmin^\ts &= \frac{32V_{max}^2}{(\zeta^\ts)^2(1-\gamma)^2} \log \frac{8\lvert S \rvert^2\lvert A \rvert^2}{\delta}.
\end{align*}
\noindent
Note that $\zeta^\ts$ and $\Nmin^\ts$ only depend on parameters of $M$ and policy performances on $\mlemdp$, which follows from \Cref{col:performance-preservation-mle} yielding $\rho(\pi_{\odot},\mlemdp)=\rho^\ts(\pi_{\odot},\mlemdpts)$.
Hence, it is not necessary to explicitly compute the transformed MLE-MDP $\mlemdpts$.

\subsection{Uncertainty in Two-Successor MDPs}\label{subsec:2s_beta}
So far, the methods we outlined relied on a bound of the $L_1$-distance between a probability vector and its estimate based on a number of samples~\cite{Weissman}. 
In this section, we outline a second method to tighten this bound for \twosMDP and how to apply it to obtain a smaller approximation error $\zeta^\beta$ for a fixed $\Nmin^\beta$.

Formally, given a \tsMDP $\Mts$ and an error tolerance $\delta$, we construct an error function $e\colon S \times A \rightarrow \Real$ that ensures with probability $1-\delta$ that $\norm{P(s,a)-\hat{P}(s,a)}_1\leq e(s,a)$ for all $(s,a)$.
To achieve this, we distribute $\delta$ uniformly over all states to obtain 
$\delta_T = \nicefrac{\delta}{|S|}$, independently ensuring that for each state-action pair $(s,a)$ the condition
$\norm{P(s,a)-\hat{P}(s,a)}_1\leq e(s,a)$ is satisfied with probability at least $1-\delta_T$.

We now fix a state-action pair $(s,a)$.
Since we are dealing with a \twosMDP, there are only two successor states, $s_1$ and $s_2$.
To bound the error function, we view each sample of action $a$ in state $s$ as a Bernoulli trial.
As shorthand notation, we define $p=P(s,a,s_1)$, and consequently we have $1-p = P(s,a,s_2)$.
Using a uniform prior over $p$ and given a data set $\data$ in which $(s,a,s_1)$ occurs $k_1$
times and $(s,a,s_2)$ occurs $k_2$ times, the posterior probability over $p$ is
given by a beta distribution with parameters $k_1{+}1$ and $k_2{+}1$, \ie, $\Pr(p \given \data) \sim B(k_1{+}1, k_2{+}1)$~\cite{Jaynes03}.
We can express the error function in terms of the probability of $p$ being contained in a given interval
$[\underline{p},\overline{p}]$ as
$
e(s,a)=\overline{p}-\underline{p}.
$

The task that remains is to find such an interval $[\underline{p},\overline{p}]$
for which we can guarantee with probability $\delta_T$ that $p$ is contained within it.
Formally, we can express this via the incomplete regularized beta function $I$,
which in turn is defined as the cumulative density function of the beta distribution $B$:
\[ 
\mathbb{P}(p\in [\underline{p},\overline{p}]) = I_{\underline{p},\overline{p}}(k_1{+}1,k_2{+}1).
\]
We show that we can define the smallest such interval in terms of the inverse incomplete beta function~\cite{Temme92}, denoted as $I_\delta^{-1}$.

\begin{lemma}\label{lemma:binomial_bound}
    Let $k \sim \mathit{Bin(n,p)}$ be a random variable according to a binomial distribution. 
    Then the smallest interval 
    $[\underline{p},\overline{p}]$ for which
    \[
    \mathbb{P}\left(p \in \left[\underline{p},\overline{p}\right]\right) \geq 1-\delta_T
    \]
    holds, has size
    \[
    \overline{p}-\underline{p} \leq 1-2I_{\nicefrac{\delta_T}{2}}^{-1}\left( \frac{n}{2}+1,\frac{n}{2}+1\right).
    \] 
\end{lemma}
Next, we show how to utilize this bound for the interval size in MDPs with arbitrary topology.
The core idea is the same as in \Cref{thm:2smdp_spi}: We transform the MDP into
a \tsMDP and apply the error bound $e(s,a)=\overline{p}-\underline{p}$ from \Cref{lemma:binomial_bound}.

\begin{theorem}[Beta-based tighter improvement guarantee]\label{thm:beta-zeta-bound}
    Let $M$ be an MDP with behavior policy $\pi_b$.
Then, $\pi_{\odot}$ is a $\zeta^\beta$-approximate safe policy
improvement over $\pi_b$ with high probability $1-\delta$, where:
\begin{align*}
\zeta^\beta = & \frac{4V_{max}}{1-\gamma} \left(1-I^{-1}_{\nicefrac{\delta_T}{2}}\left(\frac{\Nmin^\beta}{2}+1,\frac{\Nmin^\beta}{2}+1\right)\right) + \tilde{\rho},
\end{align*}
with $\delta_T = \frac{\delta}{|S|^2|A|^2}$, and $\tilde{\rho} = -\rho(\pi_{\odot},\mlemdp)+\rho(\pi_b,\mlemdp)$.
\end{theorem}

There is no closed formula to directly compute $\Nmin^\beta$ for a given $\zeta^\beta$.
However, for a given admissible performance loss $\zeta$, we can perform a binary search 
to obtain the smallest natural number $\Nmin^\beta$ such that $\zeta^\beta\leq\zeta$ given in \Cref{thm:beta-zeta-bound}.

\renewcommand\cellalign{lc}
\begin{table*}\centering
\begin{tabular}{@{}l@{\,\,\,\,\,}l@{\,\,\,\,\,}l@{}}\toprule
 Method & Admissible performance loss $\zeta$ & Number of samples $\Nmin$ \\
\midrule
  \makecell{Standard SPI \\ \cite{Petrik2016}} & $\displaystyle \zeta^{\text{SPI}} = \frac{2\gamma\Vmax}{1-\gamma} \sqrt{\frac{2}{\Nmin^{\text{SPI}}} \log \frac{2|S||A|2^{|S|}}{\delta}} $ &  $\displaystyle \Nmin^{\text{SPI}} = \frac{8 \Vmax^2}{{\zeta^{\text{SPI}}}^2 (1-\gamma)^2} \log \frac{2 |S||A|2^{|S|}}{\delta} \quad (\star)$ \\[12pt]
  \makecell{Standard SPIBB \\ \cite{DBLP:conf/icml/LarocheTC19}} & $\displaystyle \zeta^\spibb = \frac{4\Vmax}{1-\gamma} \sqrt{\frac{2}{\Nmin^\spibb} \log \frac{2|S||A|2^{|S|}}{\delta}} + \tilde{\rho}$ &  $\displaystyle \Nmin^\spibb = \frac{32 \Vmax^2}{{\zeta^\spibb}^2 (1-\gamma)^2} \log \frac{2 |S||A|2^{|S|}}{\delta}$ \\[12pt]
\makecell{Two-Successor SPIBB \\ (Theorem~\ref{thm:2smdp_spi})}& $\displaystyle \zeta^\ts = \frac{4V_{max}}{1-\gamma}\sqrt{\frac{2}{\Nmin^\ts} \log \frac{8 \lvert S \rvert^2 \lvert A \rvert^2}{\delta}} + \tilde{\rho}$ & $\displaystyle \Nmin^\ts = \frac{32V_{max}^2}{(\zeta^\ts)^2(1-\gamma)^2} \log \frac{8\lvert S \rvert^2\lvert A \rvert^2}{\delta}$ \\[12pt]
  \makecell{Inverse beta SPIBB \\ (Theorem~\ref{thm:beta-zeta-bound})} & $\displaystyle \zeta^\beta = \frac{4V_{max}}{1-\gamma}\left( 1-2I^{-1}_{\nicefrac{\delta_T}{2}}\left(\frac{\Nmin^\beta}{2}\!+\!1,\frac{\Nmin^\beta}{2}\!+\!1\right)\right) + \tilde{\rho}$ & \makecell{No closed formula available \\ (use binary search to compute)} \\[12pt]
\bottomrule
\end{tabular}
\caption{Overview of the different $\zeta$ and $\Nmin$ we obtain, where $\delta_T=\frac{\delta}{|S|^2|A|^2}$ and $\tilde{\rho} = -\rho(\pi_{\odot},\mlemdp)+\rho(\pi_b,\mlemdp)$ is the difference in performance between optimal and behavior policy on the MLE-MDP. $(\star)$ Standard SPI requires at least $\Nmin^\text{SPI}$ samples in \emph{all} state-action pairs.}
\label{tbl:zeta-Nmin-overview}
\end{table*}

\subsubsection{Comparison of Different $\Nmin$.}
In the context of SPI, finding an $\Nmin$ that is as small as possible while still guaranteeing $\zeta$-approximate improvement is the main objective.
An overview of the different $\zeta$ and $\Nmin$ that are available is given in \Cref{tbl:zeta-Nmin-overview}.

Comparing the equations for different $\Nmin$, we immediately see that $\Nmin^\ts \leq \Nmin^\spibb$ if and only if $2^{|S|} \geq 4|S||A|$. 
This means the only MDPs where standard SPIBB outperforms our \tsMDP approach are environments with a small state-space but a large action-space.

By \Cref{lemma:binomial_bound}, we have that the error term $e(s,a)$ used to compute $\zeta^\beta$ is minimal in the \tsMDP\footnote{Technically, \Cref{lemma:binomial_bound} allows for arbitrary parameters while the SPIBB algorithm only allows integers for the number of samples, and thus integer parameters in the inverse beta function, so $\zeta^\beta$ is only minimal for even $\Nmin^\beta$. 
However, we can easily adapt the equation for odd $\Nmin^{\beta}$ by replacing $\Nmin^{\beta}$ by $\Nmin^{\beta}-1$ and $\Nmin^{\beta}+1$, respectively.},
and in particular it is smaller than the error term used to compute $\zeta^{\ts}$. 
Thus we always have $\Nmin^{\beta}\leq \Nmin^{\ts}$.
In case $2^{|S|} < 4|S||A|$ it is also possible to compute both $\Nmin^\spibb$, and $\Nmin^\beta$ and simply choose the smaller one.

\begin{figure}[t]
    \centering
    \begin{subfigure}[b]{0.49\columnwidth}
        \includegraphics[width=\columnwidth]{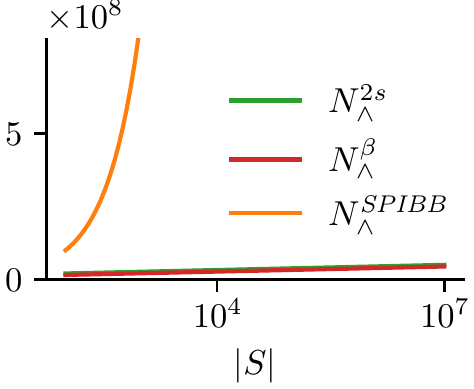}
        \caption{$\Nmin^\text{SPIBB}, \Nmin^\ts$ and $\Nmin^\beta$.}
        \label{fig:nwedge:1}
    \end{subfigure}
    \begin{subfigure}[b]{0.49\columnwidth}
        \includegraphics[width=\columnwidth]{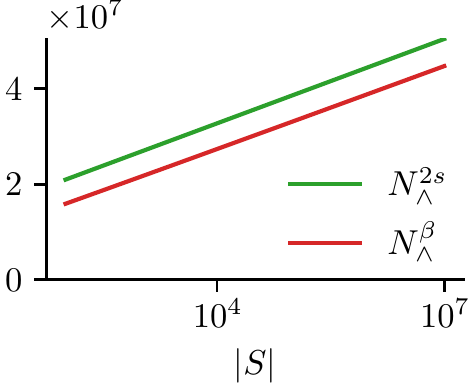}
        \caption{$\Nmin^\ts$ and $\Nmin^\beta$.}
        \label{fig:nwedge:2}
    \end{subfigure}
    \caption{Required number of samples for different $|S|$ with $|A|=4, V_{max}=1, \gamma=0.95, \delta=0.1$ and $\zeta=0.1$.}
    \label{fig:nwedges}
\end{figure}

\section{Implementation and Evaluation}\label{sec:evaluation}
We provide an evaluation\footnote{Code available at \url{https://github.com/LAVA-LAB/improved_spi}.} of our approach from two different perspectives.
First, a theoretical evaluation of how the different $\Nmin$ depend on the size of a hypothetical MDP, and second, a practical evaluation to investigate how smaller $\Nmin$ values translate to the performance of the improved policies.

\subsection{Example Comparison of Different \texorpdfstring{$\Nmin$}{N}}\label{sec:ex:example_nmin}
To render the theoretical differences between the possible $\Nmin$ discussed at the end of Section~\ref{sec:SPI_tsMDP} more tangible, we now give a concrete example.
We assume a hypothetical MDP with $|A| = 4$, $V_{max} = 1$, $\gamma = 0.95$, and SPIBB parameters $\delta = 0.1$ and $\zeta = 0.1$. 
For varying sizes of the state-space, we compute all three sample size constraints: $\Nmin^\text{SPIBB}$, $\Nmin^\ts$, and $\Nmin^\beta$.
The results are shown in \Cref{fig:nwedges}, where \Cref{fig:nwedge:1} shows the full plot and \Cref{fig:nwedge:2} provides an
excerpt to differentiate between the $\Nmin^\ts$ and $\Nmin^\beta$ plots by scaling down the $y$-axis.
Note that the $x$-axis, the number of states in our hypothetical MDP, is on a log-scale.
We see that $\Nmin^\text{SPIBB}$ grows linearly with the number of states, whereas $\Nmin^\ts$ and $\Nmin^\beta$ are logarithmic in the number of states.
Further, we note that $\Nmin^\beta$ is significantly below $\Nmin^\ts$, which follows from Lemma~\ref{lemma:2smdp_spi}.
Finally, the difference between $\Nmin^\text{SPIBB}$ and $\Nmin^\ts$ is for small MDPs of around a hundred states already a factor $10$.

\begin{figure*}[t]
    \centering
    \begin{subfigure}[t]{0.29\textwidth}
        \includegraphics[width=\columnwidth]{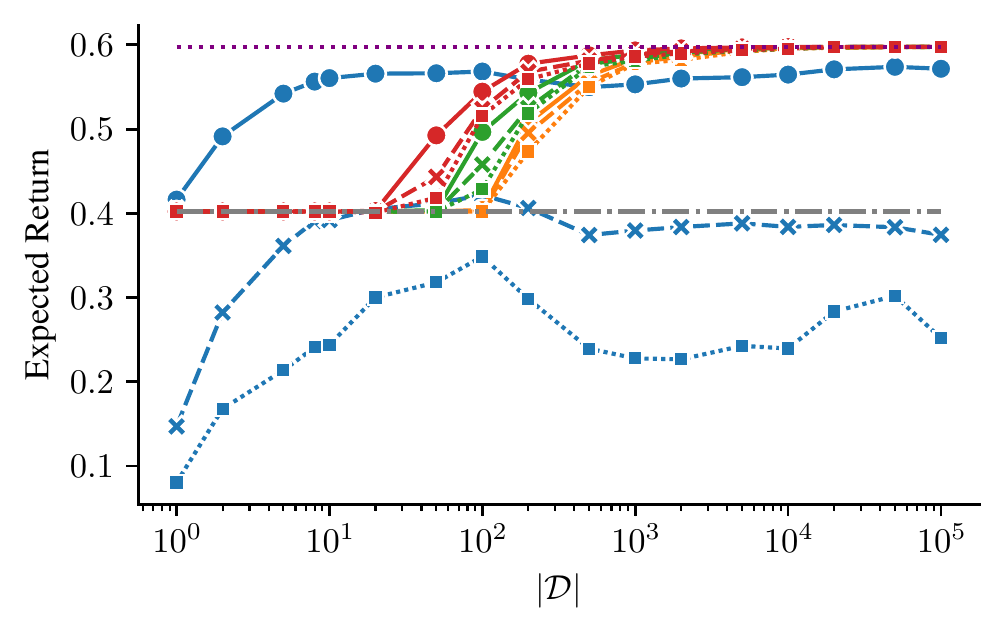}
        \caption{$\Nmin^\text{SPIBB} \!=\! 100, \Nmin^\ts \!=\! 55, \Nmin^\beta \!=\! 27$.}
    \label{fig:ex:grid100}
    \end{subfigure}
    \begin{subfigure}[t]{0.29\textwidth}
    \includegraphics[width=\columnwidth]{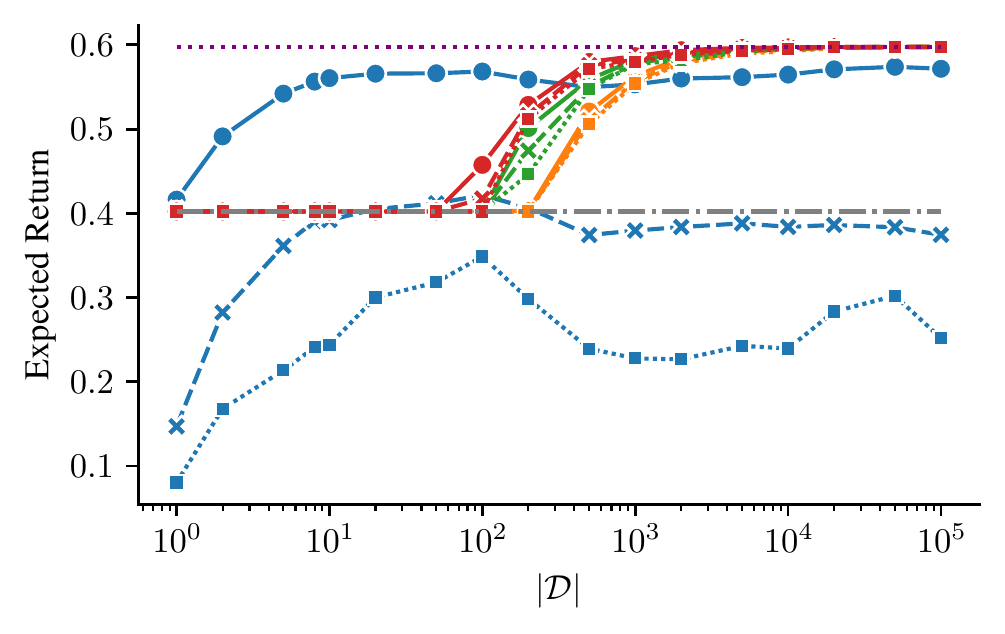}
    \caption{$\Nmin^\text{SPIBB} \!=\! 200, \Nmin^\ts \!=\! 110, \Nmin^\beta \!=\! 67$.}
    \label{fig:ex:grid200}
    \end{subfigure}
    \begin{subfigure}[t]{0.29\textwidth}
    \centering
        \includegraphics[width=\columnwidth]{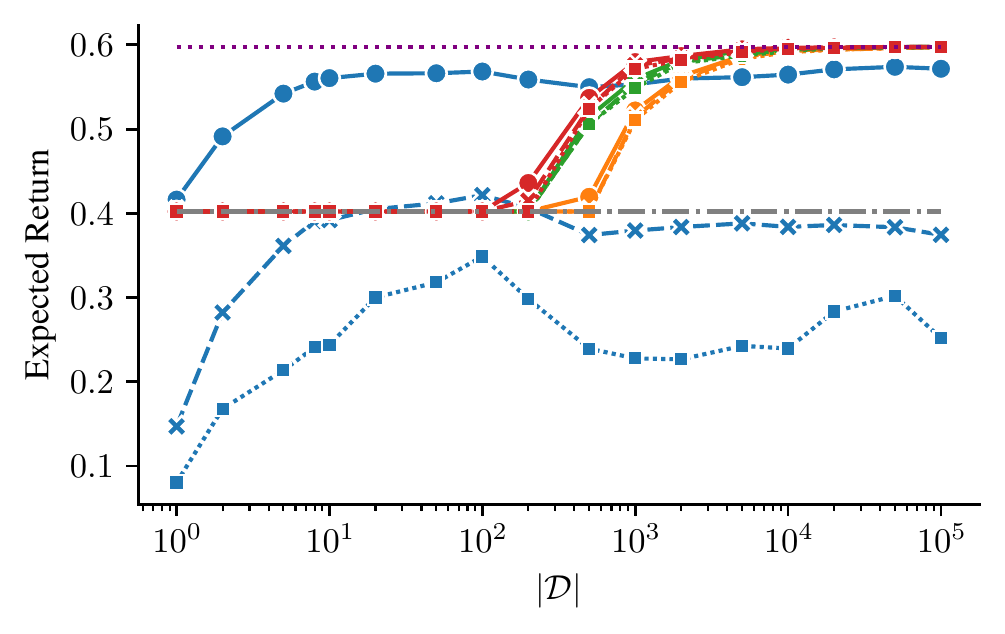}
    \caption{$\Nmin^\text{SPIBB} \!=\! 400, \Nmin^\ts \!=\! 219, \Nmin^\beta \!=\! 146$.}
    \label{fig:ex:grid400}
    \end{subfigure}
    \begin{subfigure}[t]{0.08\textwidth}
        \centering
        \includegraphics[width=\columnwidth]{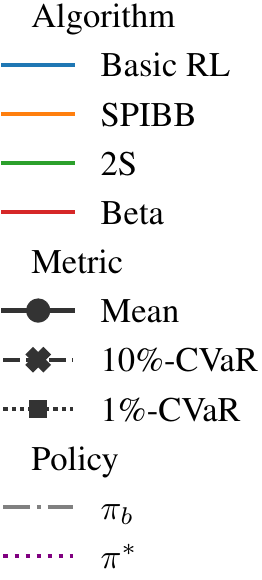}
    \end{subfigure}
    \caption{Safe policy improvement on the Gridworld environment.}
    \label{fig:ex:gridworld}
\end{figure*}

\begin{figure*}[t]
    \centering
    \begin{subfigure}[t]{0.29\textwidth}
        \includegraphics[width=\columnwidth]{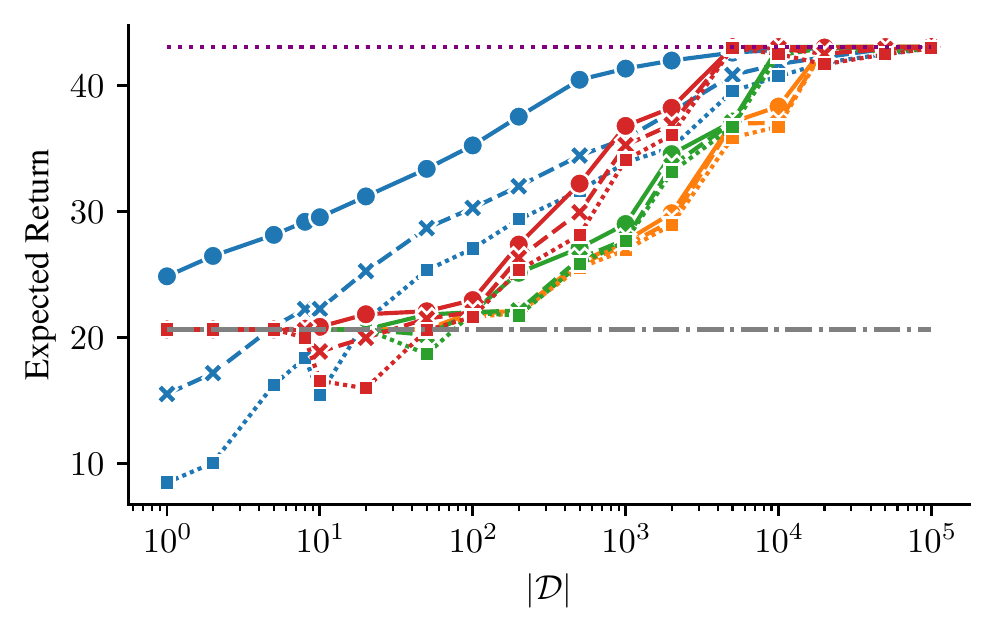}
        \caption{$\Nmin^\text{SPIBB} \!=\! 60, \Nmin^\ts \!=\! 34, \Nmin^\beta \!=\! 10$.}
    \label{fig:ex:chicken60}
    \end{subfigure}
    \begin{subfigure}[t]{0.29\textwidth}
    \includegraphics[width=\columnwidth]{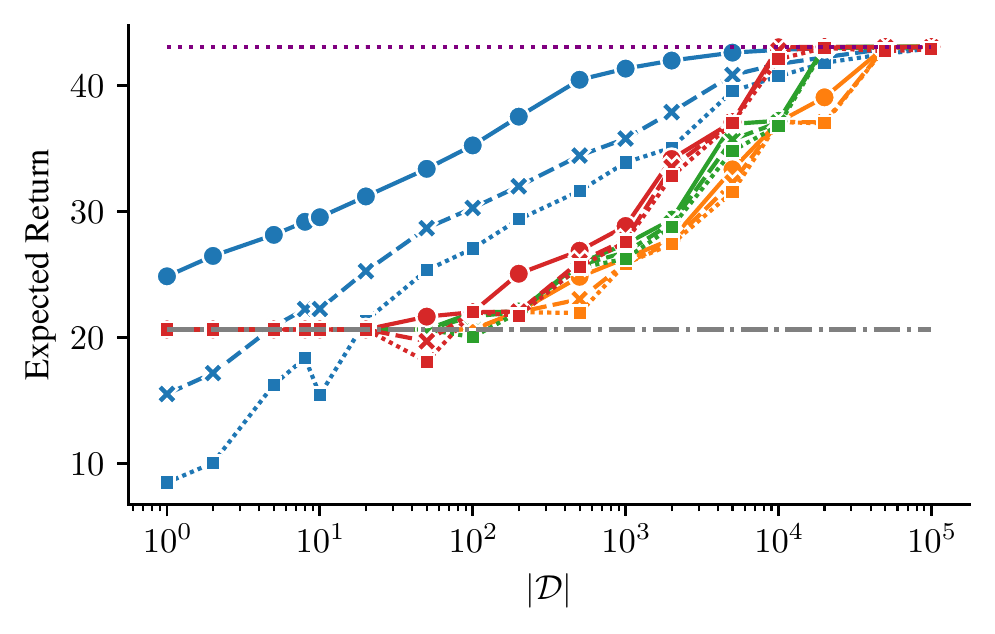}
    \caption{$\Nmin^\text{SPIBB} \!=\! 120, \Nmin^\ts \!=\! 67, \Nmin^\beta \!=\! 36$.}
    \label{fig:ex:chicken120}
    \end{subfigure}
    \begin{subfigure}[t]{0.29\textwidth}
    \centering
        \includegraphics[width=\columnwidth]{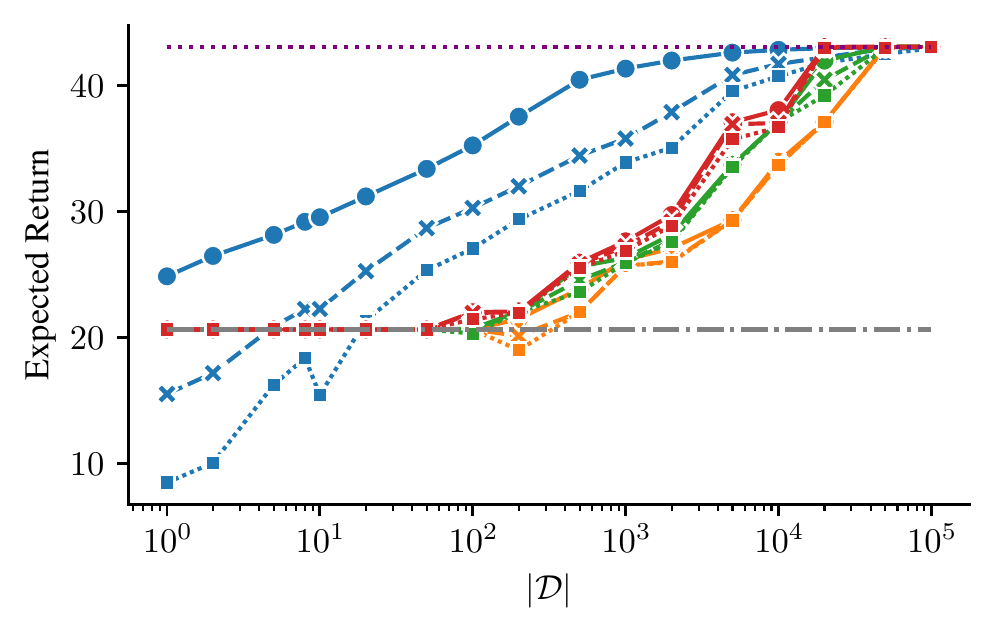}
    \caption{$\Nmin^\text{SPIBB} \!=\! 180, \Nmin^\ts \!=\! 101, \Nmin^\beta \!=\! 61$.}
    \label{fig:ex:chicken180}
    \end{subfigure}
    \begin{subfigure}[t]{0.08\textwidth}
        \centering
        \includegraphics[width=\columnwidth]{imgs/legend.pdf}
    \end{subfigure}
    \caption{Safe policy improvement on the Wet Chicken environment.}
    \label{fig:ex:chicken}
\end{figure*}

\begin{figure*}[t]
    \centering
    \begin{subfigure}[t]{0.29\textwidth}
        \includegraphics[width=\columnwidth]{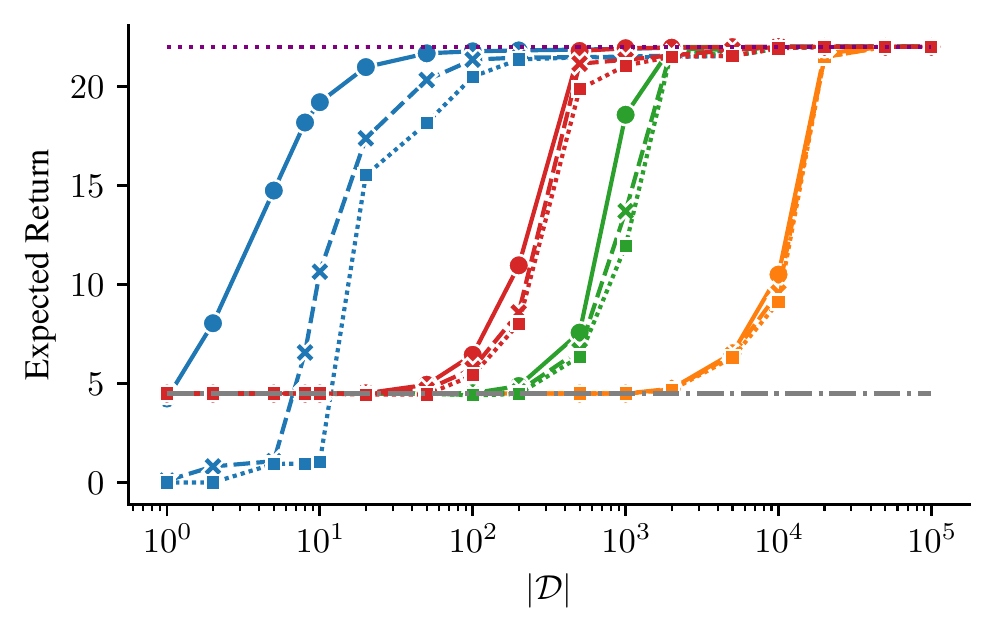}
        \caption{$\Nmin^\text{SPIBB} \!=\! 600, \Nmin^\ts \!=\! 43, \Nmin^\beta \!=\! 12$.}
    \label{fig:ex:resource600}
    \end{subfigure}
    \begin{subfigure}[t]{0.29\textwidth}
    \includegraphics[width=\columnwidth]{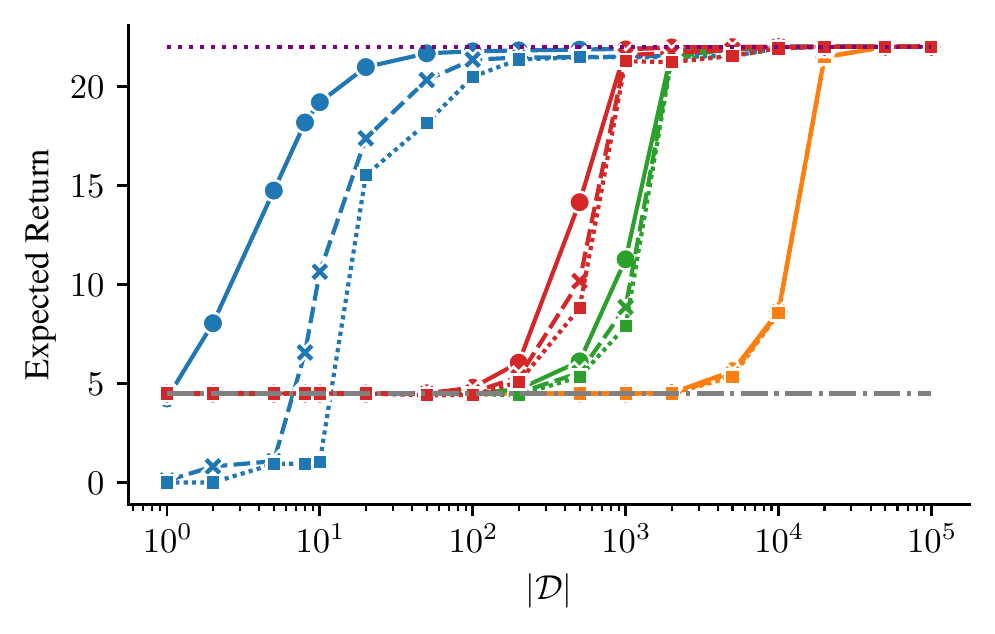}
    \caption{$\Nmin^\text{SPIBB} \!=\! 800, \Nmin^\ts \!=\! 57, \Nmin^\beta \!=\! 25$.}
    \label{fig:ex:resource800}
    \end{subfigure}
    \begin{subfigure}[t]{0.29\textwidth}
    \centering
        \includegraphics[width=\columnwidth]{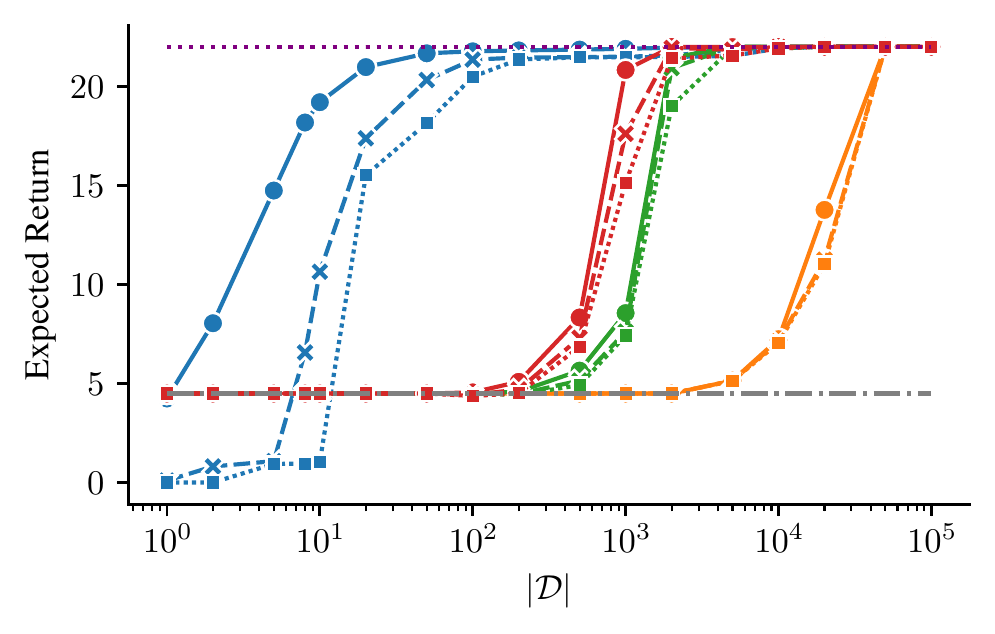}
    \caption{$\Nmin^\text{SPIBB} \!=\! 1000, \Nmin^\ts \!=\! 71, \Nmin^\beta \!=\! 37$.}
    \label{fig:ex:resource1000}
    \end{subfigure}
    \begin{subfigure}[t]{0.08\textwidth}
        \centering
        \includegraphics[width=\columnwidth]{imgs/legend.pdf}
    \end{subfigure}
    \caption{Safe policy improvement on the Resource Gathering environment.}
    \label{fig:ex:resource}
\end{figure*}

\subsubsection{Discussion. } 
While we show that a significant reduction of the required number of samples per state-action pair $\Nmin$ is possible via our two approaches, we note that even for 
small MDPs (\eg, $|S| = 100$) we still need over $10$ million samples per state-action pair to guarantee that an improved policy is safe \emph{w.r.t.} the behavior policy.
That is, with probability $1-\delta = 0.9$, an improved policy will have an admissible performance loss of at most $\zeta = 0.1$, which is infeasible in practice.
Nevertheless, a practical evaluation of our approaches is possible taking on a different perspective, which we address in the next section.

\subsection{Evaluation in SPIBB}
We integrate our novel results for computing $\zeta^\ts, \zeta^\beta, \Nmin^\ts$, and $\Nmin^\beta$ into the implementation of SPIBB~\cite{DBLP:conf/icml/LarocheTC19}. 
\paragraph{Benchmarks.}
We consider two standard benchmarks used in SPI and one other well-known MDP: 
the $25$-state \emph{Gridworld} proposed by~\citet{DBLP:conf/icml/LarocheTC19}, the $25$-state \emph{Wet Chicken} benchmark~\cite{DBLP:conf/icann/HansU09}, which was used to evaluate SPI approaches by~\citet{DBLP:conf/icaart/SchollDOU22}, and a $376$-state instance of \emph{Resource Gathering} proposed by~\citet{DBLP:conf/icml/BarrettN08}.

\paragraph{Behavior policy.}
For the Gridworld, we use the same behavior policy as~\cite{DBLP:conf/icml/LarocheTC19}.
For the Wet Chicken environment, we use Q-Learning with a softmax function to derive a behavior policy.
The behavior policy of Resource Gathering was derived from the optimal policy by selecting each non-optimal action with a probability of \texttt{1e-5}. 

\paragraph{Methodology. } 
Recall that in the standard SPIBB approach, $\Nmin$ is used as a hyperparameter, since the actual $\Nmin$ for reasonable $\delta$ and $\zeta$ are infeasible.
While our methods improve significantly on $\Nmin$, the values we obtain are still infeasible in practice, as discussed in Section~\ref{sec:ex:example_nmin}.
We still use $\Nmin^\text{SPIBB}$ as a hyperparameter, and 
then run the SPIBB algorithm and compute the resulting $\zeta^\text{SPIBB}$.
This $\zeta^\text{SPIBB}$ is consequently used to compute the values $\Nmin^\ts$ and $\Nmin^\beta$ that ensure the same performance loss.
We then run SPIBB again with these two values for $\Nmin$.
As seen in the previous experiment, and detailed at the end of Section~\ref{sec:SPI_tsMDP}, for most MDPs -- including our examples -- we have $\Nmin^\beta \leq \Nmin^\ts \leq \Nmin^\text{SPIBB}$ for a fixed $\zeta$.

\paragraph{Evaluation metrics.}
For each data set size, we repeat each experiment $1000$ times and report the mean performance of the learned policy, as well as the $10\%$ and $1\%$ conditional value at risk (CVaR) values~\cite{rockafellar2000optimization}, indicating the mean performance of the worst $10\%$ and $1\%$ runs. 
To give a complete picture, we also include the performance of basic RL (dynamic programming on the MLE-MDP~\cite{DBLP:books/lib/SuttonB98}), the behavior policy $\pi_b$, and the optimal policy $\pi^*$ of the underlying MDP.

\subsubsection{Results.}
We present the results for the Gridworld, Wet Chicken, and Resource Gathering environments for three different hyperparameters $\Nmin^\text{SPIBB}$ in Figures~\ref{fig:ex:gridworld},~\ref{fig:ex:chicken}, and~\ref{fig:ex:resource}, respectively.
In all instances, we see similar and improved behaviors as we presumed by sharpening the sampling bounds with our new approaches.
Smaller values for $\Nmin$ typically require smaller data sets for a policy to start improving, and this is precisely what our methods set out to do.
In particular, we note that our methods (2S and Beta) are quicker to converge to an optimal policy than standard SPIBB.
Beta is, as expected, the fastest, and starts to improve over the behavior policy for data sets about half the size compared to SPIBB in the Gridworld.
Further, while theoretically, the factor between the different $\Nmin$ does not directly translate to the whole data set size, we see that in practice on all three benchmarks this is roughly the case.
Finally, we note that Basic RL is unreliable compared to the SPI methods, as seen by the CVaR values being significantly below the baseline performance for several data set sizes in all three environments.
This is as expected and in accordance with well-established results.

\section{Related Work}\label{sec:related-work}
A variant of our transformation from MDP to \tsMDP was introduced by~\citeauthor{MayMun23}~\shortcite{MayMun23}, utilizing binary trees built from auxiliary states as gadgets.
Similar to our construction, \citeauthor{DBLP:conf/uai/Junges0WQWK018}~\shortcite{DBLP:conf/uai/Junges0WQWK018} transform a partially observable MDP (POMDP)~\cite{DBLP:journals/ai/KaelblingLC98,DBLP:books/sp/12/Spaan12} into a \emph{simple POMDP}, where each state has either one action choice, and an arbitrary number of successor states, or where there are multiple actions available but each action has a single successor state.
The same transformation was applied to \emph{uncertain} POMDPs~\cite{DBLP:conf/aaai/Cubuktepe0JMST21}.

Besides the main approaches to SPI mentioned in Section~\ref{sec:SPI}, there are a number of other noteworthy works in this area.
SPIBB has been extended to \emph{soft baseline bootstrapping} in~\cite{DBLP:conf/pkdd/NadjahiLC19}, where instead of either following the behavior policy or the optimal policy in the MLE-MDP in a state-action pair, randomization between the two is applied.
However, the theoretical guarantees of this approach rely on an assumption that rarely holds~\cite{DBLP:conf/icaart/SchollDOU22}.

Incorporating structural knowledge of the environment has been shown to improve the sample complexity of SPI algorithms~\cite{DBLP:conf/aaai/SimaoS19,DBLP:conf/ijcai/SimaoS19}.
It is also possible to deploy the SPIBB algorithm in problems with large state space using MCTS \cite{Castellini2023}.
For a more detailed overview of SPI approaches and an empirical comparison between them, see~\cite{DBLP:conf/icaart/SchollDOU22}.
For an overview of how these algorithms scale in the number of states, we refer to~\cite{brandfonbrener2022incorporating}.

Other related work investigated how to relax some of the assumptions SPI methods make.
In~\cite{DBLP:conf/atal/SimaoLC20}, a method for estimating the behavior policy is introduced, relaxing the need to know this policy.
Finally, a number of recent works extend the scope and relax common assumptions by introducing SPI in problems with partial observability~\cite{Simao2023SPIPOMDP}, non-stationary dynamics~\cite{DBLP:conf/nips/ChandakJTWT20}, and multiple objectives~\cite{DBLP:conf/nips/SatijaTPL21}.

Finally, we note that SPI is a specific \emph{offline} RL problem~\cite{DBLP:journals/corr/abs-2005-01643} and
there have been significant advances in the general offline setting recently~\cite{DBLP:conf/nips/KidambiRNJ20,Yu2020,DBLP:conf/nips/KumarZTL20,Smit21IJCAIWS,DBLP:conf/nips/YuKRRLF21,DBLP:journals/corr/abs-2204-12581}. 
While these approaches may be applicable to high dimensional problems such as control tasks and problems with large observation space \cite{fu2020d4rl}, they often ignore the inherent reliability aspect of improving over a baseline policy, as SPI algorithms do.
Nevertheless, it remains a challenge to bring SPI algorithms to high-dimensional problems.
\section{Conclusion}\label{sec:conclusion}
We presented a new approach to safe policy improvement that reduces the required size of data sets significantly.
We derived new performance guarantees and applied them to state-of-the-art approaches such as SPIBB.
Specifically, we introduced a novel transformation to the underlying MDP model that limits the branching factor, and provided two new ways of computing the admissible performance loss $\zeta$ and the 
sample size constraint $\Nmin$, both exploiting the limited branching factor in SPI(BB).
This improves the overall performance of SPI algorithms, leading to more 
efficient use of a given data set.

\section*{Acknowledgments}
  The authors were partially supported by the DFG through
    the Cluster of Excellence EXC 2050/1 (CeTI, project ID 390696704, as part of Germany's 
 	Excellence Strategy),
    the TRR 248 (see \url{https://perspicuous-computing.science}, project ID 389792660),
    the NWO grants OCENW.KLEIN.187 (Provably Correct Policies for Uncertain Partially Observable Markov Decision Processes) and NWA.1160.18.238 (PrimaVera), and 
    the ERC Starting Grant 101077178 (DEUCE).

\bibliographystyle{named}
\bibliography{main}

\newpage
\onecolumn
\appendix

\section{Proofs}
\label{ap:proofs}

\begin{proof}[Proof Theorem~\ref{thm:mdp_transformation}]
\hspace{0cm}\\
Fix a state $s\in S$ and action $a\in A$ that is enabled in $s$ and assume $\Post_M(s,a) = \{s_1,\dots,s_k\}$. We define $p_i = P(s_i \given s,a)$ for all $i=1,\dots,k$.

By definition, for each $s_i \in \Post_M(s,a)$ with $i<k$ a path $\langle s,a, x_2,\tau, \dots, x_i,\tau, s_i \rangle$ and has positive probability. For $i=k$ the path is $\langle s,a, x_2,\tau, \dots, x_{k-1},\tau, s_k\rangle$ instead.
Note that there is exactly one incoming transition for each auxiliary state $x_i \in \Saux^{s,a}$ and each $s_i \in \Post_M(s,a)$ has exactly one predecessor in $\Saux^{s,a}$.
Further, $\Post_\Mts(s,a) \subseteq \Saux^{s,a} \cup \Post_M(s,a)$ and also $\Post_\Mts(x,\tau) \subseteq \Saux^{s,a} \cup \Post_M(s,a)$ for all $x\in \Saux^{s,a}$, \ie, any path $\langle s,a,\dots \rangle$ must stay in the set $\Saux^{s,a}$ until it reaches a state in $\Post_M(s,a)$. Thus the path from $s$ to $s_i$ must be unique.

We now show that the probabilities pre- and post-transformation match, \ie,
$\prob_M(\langle s, a, s_i \rangle) = \prob_{\Mts}(\langle s, a, x_2, \tau, \dots, x_i, \tau, s_i \rangle)$.
For $s_i=s_1$ the statement holds by definition. Now, consider the case $2\leq i \leq k-1$.
Then we can compute the probability of the path as 

\begin{align*}
    \prob_{\Mts}(\langle s, a, x_2, \tau, \dots, x_i, \tau, s_i \rangle) &= 
    P^\ts(x_2 \given s,a) \left( \prod_{j=2}^{i-1} P^\ts(x_{j+1} \given x_j,\tau) \right) P^\ts(s_i \given x_i,\tau)  \\
    &= (1-p_1) \left( \prod_{j=2}^{i-1} 1-\frac{p_j}{1-(p_1+\dots +p_{j-1})} \right) \frac{p_i}{1-(p_1+\dots +p_{i-1})}  \\
    &= (1-p_1) \left( \prod_{j=2}^{i-1} \frac{1-(p_1+\dots +p_{j})}{1-(p_1+\dots +p_{j-1})} \right) \frac{p_i}{1-(p_1+\dots +p_{i-1})}  \\
    &= p_i \\
    &= \prob_M(\langle s, a, s_i \rangle).
\end{align*}
For the case $i=k$ the proof is analogous to the case $i=k-1$, replacing the last factor by $P^\ts(s_k \given x_{k-1},\tau)$.
\end{proof}

\begin{proof}[Proof Corollary~\ref{col:performance-preservation}]
\hspace{0cm}\\
Recall that the auxiliary states can only choose action $\tau$ which does not yield any reward.
Further, since no discount is applied in auxiliary states, for all states $s \in S$ we have
\begin{align*}
V_\Mts^\pits(s) \ &= \ 
		\sum\limits_{a \in A} \pits(a \given s) \Big( R^\ts(s,a) + \gamma^\ts(s) \sum\limits_{s' \in S} P^\ts(s' \given s, a) V_\Mts^\pits(s') \Big) \\
  &= \ 
		\sum\limits_{a \in A} \pi(a \given s) \Big( R(s,a) + \gamma \sum\limits_{s' \in S} \prob_{\Mts}(s_i \given s, a, x_2, \tau, \dots, x_i, \tau) V_\Mts^\pits(s') \Big)    \\
  &= \
        \sum\limits_{a \in A} \pi(a \given s) \Big( R(s,a) + \gamma \sum\limits_{s' \in S} P(s,a) V_\Mts^\pits(s') \Big).
\end{align*}
This is exactly the defining Bellman equation for $V_m^\pi$, \ie, by definition, we have
$$V_m^\pi(s)=\sum\limits_{a \in A} \pi(a \given s) \Big( R(s,a) + \gamma \sum\limits_{s' \in S} P(s,a) V_M^\pi(s') \Big).  $$

As the Bellman equation has a unique fixed point for a discount factor $\gamma<1$ (see e.g. \cite{DBLP:books/lib/SuttonB98}), we must have $V_m^\pi = V_\Mts^\pits(s)$ for all $s\in S$.
In particular, this holds for $s=\iota$, and thus $\rho(M,\pi)=\rho(\Mts,\pits)$
\end{proof}

\begin{proof}[Proof Theorem~\ref{thm:data_transformation}]
\hspace{0cm}\\
The proof is similar to the proof of \Cref{thm:mdp_transformation}.

Again, fix a state $s\in S$ and action $a\in A$ that is enabled in $s$ and assume $\Post_{\mlemdp}(s,a) = \{s_1,\dots,s_k\}$. For the case $k\leq 2$ the statement trivially holds as $\cnt_\data(s,a,\cdot)=\cnt_\datats(s,a,\cdot)$. We now consider the case $k>2$.

As we have $\tilde{P}^\ts(s' \given s,a)>0$ if and only if $\cnt_\datats(s,a,s')>0$ for all $s,s'\in S\cup\Saux$, and for each $s' \in \Saux^{s,a} \cup \Post_{\mlemdp}(s,a)$
by definition there is exactly one state $s \in \Saux^{s,a} \cup \{s\}$ for which $\cnt_\datats(s,a,s')>0$, there for each $i\in\{1,\dots,k\}$ must be exactly one unique path $\langle s, a, x_2, \tau, \dots, x_i, \tau, s_i \rangle$ in $\mlemdpts$.

We now show that
$\prob_{\mlemdp}(\langle s, a, s_i \rangle) = \prob_{\mlemdpts}(\langle s, a, x_2, \tau, \dots, x_i, \tau, s_i \rangle )$.
For $s_i=s_1$ the statement holds by definition. Now consider the case $2\leq i \leq k-1$.
Then we can compute the probability of the path as 
\begin{align*}
    \prob_{\mlemdpts}(\langle s, a, x_2, \tau, \dots, x_i, \tau, s_i \rangle) &= 
    P_\mlemdpts(x_2 \given s,a) \left( \prod_{j=2}^{i-1} P_\mlemdpts(x_{j+1} \given x_j,\tau) \right) P_\mlemdpts(s_i \given x_i,\tau)  \\
    &= \frac{\cnt_\datats(s,a,x_2)}{\sum_{s'}\cnt_\datats(s,a,s')} \left( \prod_{j=2}^{i-1} \frac{\cnt_\datats(x_i,\tau,x_{j+1})}{\sum_{s'}\cnt_\datats(x_j,\tau,s')} \right) \frac{\cnt_\datats(x_i,\tau,s_i)}{\sum_{s'}\cnt_\datats(x_i,\tau,s')}  \\
    &= \frac{\cnt_\datats(s,a,x_2)}{\cnt_\datats(s,a,s_1)+\cnt_\datats(s,a,x_2)} \left( \prod_{j=2}^{i-1} \frac{\cnt_\datats(x_i,\tau,x_{j+1})}{\cnt_\datats(x_j,\tau,s_j)+\cnt_\datats(x_j,\tau,x_{j+1})} \right) \\ 
    & \phantom{==} \frac{\cnt_\datats(x_i,\tau,s_i)}{\cnt_\datats(x_i,\tau,s_i)+\cnt_\datats(x_i,\tau,x_{i+1})}  \\
    &= \frac{\sum_{m=2}^k\cnt_\data(s,a,s_m)}{\cnt_\data(s,a,s_1)+\sum_{m=2}^k\cnt_\data(s,a,s_m)} \left( \prod_{j=2}^{i-1} \frac{\sum_{m=j+1}^k\cnt_\data(s,a,s_m)}{\cnt_\data(s,a,s_j)+\sum_{m=j+1}^k\cnt_\data(s,a,s_m)} \right) \\ 
    & \phantom{==} \frac{\cnt_\data(s,a,s_i)}{\cnt_\data(s,a,s_i)+\sum_{m=i+1}^k\cnt_\data(s,a,s_m)}  \\
    &= \frac{\sum_{m=2}^k\cnt_\data(s,a,s_m)}{\sum_{m=1}^k\cnt_\data(s,a,s_m)} \left( \prod_{j=2}^{i-1} \frac{\sum_{m=j+1}^k\cnt_\data(s,a,s_m)}{\sum_{m=j}^k\cnt_\data(s,a,s_m)} \right) \frac{\cnt_\data(s,a,s_i)}{\sum_{m=i}^k\cnt_\data(s,a,s_m)}  \\
    &= \frac{\cnt_\data(s,a,s_i)}{\sum_{m=1}^k\cnt_\data(s,a,s_m)} \\
    &= \frac{\cnt_\data(s,a,s_i)}{\sum_{s'}\cnt_\data(s,a,s')} \\
    &= \prob_\mlemdp(s_i \given s, a).
\end{align*}
The case where $i=k$ is analogous to the case $i=k-1$.
\end{proof}

\begin{proof}[Proof Corollary~\ref{col:performance-preservation-mle}]
\hspace{0cm}\\
Using \Cref{thm:data_transformation}, the proof is analogous to the proof of \Cref{col:performance-preservation}.
\end{proof}

\begin{proof}[Proof Lemma~\ref{lemma:2smdp_spi}]
\hspace{0cm}\\
The proof follows a similar argumentation as \cite[Theorem 2]{DBLP:conf/icml/LarocheTC19}. First, note that we cannot directly apply the cited
theorem as we deal with a \twosMDP for which $\rho(\cdot,\cdot)$ is defined in a different manner. More precisely, the discount rate $\gamma$ is not constant.
We now show that the results can still be applied in our setting by adapting the proof of \cite[Theorem 2]{DBLP:conf/icml/LarocheTC19}.

Let $\Mts = (\Smain \cup \Saux, A \cup \{\tau\}, P^\ts, R^\ts, \gamma)$ be the given \twosMDP, $\cD$ a data-set of trajectories over $\Mts$ and 
$\Nmin \in \Nat$ a given parameter.
The set of bootstrapped state-action pairs is defined as $\cU=\{(s,a)\mid \#_{\datats}(s,a)<\Nmin\} \cup \{(s,\tau)\mid s\in\Smain\}$.
Note that actions in auxiliary states are always bootstrapped.
The behavior policy $\pi_b$ is decomposed into $\tilde{\pi}_b$, for bootstrapped actions, and $\dot{\pi}_b$, for non-bootstrapped actions, i.e.,
\[
\tilde{\pi}_b = \begin{cases}
    \pi_b(a\given s) & (a\given s)\in\cU \\
    0 & \text{else}
\end{cases}
\]
and $\dot{\pi}_b(a\given s) = \pi_b(a\given s) - \tilde{\pi}_b(a\given s)$.

We now transform $\Mts$ into a its corresponding bootstrapped semi-MDP \cite{DBLP:books/lib/SuttonB98} counterpart $\ddot{M}$.
Precisely, $\ddot{M}=(\Smain \cup \Saux,\Omega_A, \ddot{P}, \ddot{R}, \Gamma)$ with $\Omega_A=\{\omega_a\}_{a\in A}$ where for each $a\in A$
\[
\omega_a = \langle I_a, a:\pi_b, \beta(s) \rangle \begin{cases} I_a = \{ s \mid (s,a) \not\in \cU \} \\
a:\pi_b \dots \text{choose $a$, then follow $\pi_b$} \\
\beta(s) = \sum_{a'\in A\cup\{\tau\}}{\dot{\pi}_b(s,a')}
\end{cases}
\]
and
\[
\Gamma(s,a)=\begin{cases}
    \gamma & s\in\Smain \\
    1 & s\in\Saux
\end{cases}
\]
$P$ and $R$ are naturally extended to options, i.e., $\ddot{P}(s,\omega_a)=P^{2s}(s,a)$ and $\ddot{R}(s,\omega_a)=R^{2s}(s,a)$.
In the same fashion we transform the MLE MDP $\mlemdpts$ into its bootstrapped counterpart $\tilde{\ddot{M}}^{2s}=(\Smain \cup \Saux,\Omega_A, \tilde{\ddot{P}}, \ddot{R}, \Gamma)$.

Using Theorem 2.1 from \cite{Weissman} on all $(s,a)\not\in\cU$ we obtain that for all $a\in A$ and $s\in I_a$, with probability at least $1-\delta$ we have
\begin{align*}
    \norm{\gamma \ddot{P}(s,a) - \gamma\tilde{\ddot{P}}(s,a)}_1 & \leq \sqrt{\frac{2}{\min_{a\in A,s\in I_a} \cnt_{\datats}(s,a)} \log \frac{2 |S||A|2^{\max_{a\in A,s\in I_a}|\Post(s,a)|}}{\delta} } \\
    & \leq \sqrt{\frac{2}{\Nmin} \log \frac{8 |S||A|}{\delta}  }
\end{align*}

This means we can apply Lemma 1 from \cite{DBLP:conf/icml/LarocheTC19} to $\Mts$ and $\mlemdpts$ for arbitrary bootstrapped policies to obtain
\begin{align*}
    |\rho(\pi^{\ts}_{\odot}, \Mts)-\rho(\pi^{\ts}_{\odot}, \mlemdpts)| &\leq \frac{2V_{max}}{1-\gamma} \sqrt{\frac{2}{\Nmin} \log \frac{8 |S||A|}{\delta}} \\
    |\rho(\pi^\ts_b, \Mts)-\rho(\pi^\ts_b, \mlemdpts)| &\leq \frac{2V_{max}}{1-\gamma}\sqrt{\frac{2}{\Nmin} \log \frac{8 |S||A|}{\delta}}
\end{align*}

Combining these inequalities we obtain the final result

\begin{align*}
    \zeta^{2s} &= \rho(\pi^\ts_b, \Mts) - \rho(\pi^{\ts}_{\odot}, \Mts)\\
    &\leq \frac{4V_{max}}{1-\gamma}\sqrt{\frac{2}{\Nmin} \log \frac{8 |S||A|}{\delta}} - \rho(\pi^{\ts}_{\odot}, \mlemdpts) + \rho(\pi^\ts_b, \mlemdpts). \qedhere
\end{align*}

\end{proof}

\begin{proof}[Proof Theorem~\ref{thm:2smdp_spi}]
\hspace{0cm}\\

\begin{align*}
\rho(\pi_{\odot},M) &= \rho(\pi^{\ts}_{\odot},\Mts) \\
&\geq \rho(\pi_b^\ts,\Mts)-\frac{4V_{max}}{1-\gamma}\sqrt{\frac{2}{\Nmin} \log \frac{8 \lvert S \rvert^2 \lvert A \rvert^2}{\delta}} -\rho(\pi^{\ts}_{\odot},\mlemdpts)+\rho(\pi_b^\ts,\mlemdpts)  \\
&= \rho(\pi_b,M)-\zeta^\ts
\end{align*}

Both equalities follow by applying \Cref{thm:mdp_transformation} and \Cref{thm:data_transformation}. The inequality is obtained by applying 
\Cref{lemma:2smdp_spi} to $\Mts$ where the size of the state space of $\Mts$ is bounded by $|\Smain\cup\Saux|\leq|S|^2|A|$ (cf. \Cref{subsec:mdp_transformation}).
\end{proof}

\paragraph{$k$-successor MDPs.}
\phantomsection
\label{par:k_suc_mdp}
We saw that the $\Nmin^{\text{SPIBB}}$ grows linearly in $|S|$ whereas $\Nmin^{\ts}$
grows logarithmically in $|S|$. The contributing factor to this was that for any
state-action pair $(s,a)$ the $L_1$-norm between the true  successor distribution $P(s,a)$
and its maximum likelihood estimate can only be bounded linearly in terms of the branching
factor $k=|\Post(s,a)|$ when applying the bound obtained by \cite{Weissman}.
Hence, the idea behind our 2sMDP transformation was to bound the branching factor by a 
constant. To achieve this, we necessarily needed to introcude auxiliary state, essentially 
creating a trade-off between branching factor and size of the state space.
One question that may arise is whether it may be beneficial to allow for a larger 
branching factor $k$ than $2$, possibly harvesting the advantage of having to introduce 
less auxiliary states. In \Cref{subsec:mdp_transformation} we hint at the fact that for
SPI algorithms, $k=2$ is indeed optimal. We now outline why this is the case.

First, notice that we can easily adapt the transformation outlined in 
\Cref{subsec:mdp_transformation} towards $k$-successor MDPs by structuring the auxiliary 
nodes in a tree with branching factor $2$ rather than in a binary tree, resulting in up to 
$\nicefrac{|S|}{k-1}$ auxiliary states in each tree. Thus, when transforming an arbitrary 
MDP into a $k$-successor MDP, we can bound the $L_1$-error in the same fashion as in
\Cref{thm:2smdp_spi} to obtain

\[
N_{\wedge}^{ks} = \frac{32V_{max}^2}{\zeta^2(1-\gamma)^2} \log \frac{2\lvert S \rvert^2\lvert A \rvert^2 2^k}{(k-1)\delta}
\]

As all terms are positive his expression is minimal if and only if $\frac{2^k}{k-1}$ is 
minimal, which is the case for $k=2$ and $k=3$. Hence, 2sMDP are optimal for SPIBB
when utilizing the $L_1$-norm bound by \cite{Weissman}.

The main reason why we choose a branching factor of $k=2$ over $k=3$ is that for $k=2$ we 
can give even tighter bounds on the $L_1$-norm by computing integrals over the pdf of the
transition probabilities that is given by a beta distribution as described in \Cref{subsec:2s_beta}. Next, we provide proofs for the Lemma and Theorem in that section.

\begin{proof}[Proof Lemma~\ref{lemma:binomial_bound}]
\hspace{0cm}\\

We first show that for $k=\frac{n}{2}$ the statement holds.

Let $B(\cdot,\cdot)$ denote the beta function and $f_B(\cdot;a,b)$ the probability density function of the Beta distribution with parameters $a$ and $b$.

Consider the interval $[\underline{p},\overline{p}]=[\frac{1}{2}-h,\frac{1}{2}+h]$ with $h=\frac{1}{2}\left(1-2I_{\nicefrac{\delta_T}{2}}^{-1}\left( \frac{n}{2}+1,\frac{n}{2}+1\right)\right)$.
The interval clearly has size $\overline{p}-\underline{p}=1-2I_{\nicefrac{\delta_T}{2}}^{-1}\left( \frac{n}{2}+1,\frac{n}{2}+1\right)$.
Now we show that it contains $p$ with probability $1-\delta_T$.
\begin{align}
\mathbb{P}\left(p \in \left[\underline{p},\overline{p}\right]\right) & = \int_{\frac{1}{2}-h}^{\frac{1}{2}+h} f_B(u,\frac{n}{2}+1,\frac{n}{2}+1) \label{line:2}\\
& = \int_{\frac{1}{2}-h}^{\frac{1}{2}+h} \frac{u^{\frac{n}{2}}\left(1-u\right)^{\frac{n}{2}}}{B\left(\frac{n}{2}+1,\frac{n}{2}+1\right)} du \label{line:3}\\
& = I_{\frac{1}{2}+h}\left(\frac{n}{2}+1,\frac{n}{2}+1\right) - I_{\frac{1}{2}-h}\left(\frac{n}{2}+1,\frac{n}{2}+1\right) \label{line:4}\\
& = 1-2\cdot I_{\frac{1}{2}-h}\left(\frac{n}{2}+1,\frac{n}{2}+1\right) \label{line:5}\\
& = 1-2I_{I_{\nicefrac{\delta_T}{2}}^{-1}\left( \frac{n}{2}+1,\frac{n}{2}+1\right)}\left(\frac{n}{2}+1,\frac{n}{2}+1\right) \label{line:6}\\
& = 1-2\frac{\delta_T}{2} \label{line:7}\\
& = 1-\delta_T
\end{align}
Note that \Cref{line:2} assumes a unform prior. \Cref{line:4} is obtained by definition and \Cref{line:5} by using the symmetry of $I(a,a)$.

Due to symmetry of $f_B(\cdot,\frac{n}{2},\frac{n}{2})$ and its monotonicity on the intervals
$[0,\frac{1}{2})$ and $(\frac{1}{2},1]$ we have for all $r\in[0,\frac{1}{2}-h]\cup[\frac{1}{2}+h,1]$ and $s\in[\frac{1}{2}-h,\frac{1}{2}+h]$ that 
$$f_B(r,\frac{n}{2},\frac{n}{2}) \leq f_B(\frac{1}{2}-h,\frac{n}{2},\frac{n}{2}) = f_B(\frac{1}{2}+h,\frac{n}{2},\frac{n}{2}) \leq f_B(s,\frac{n}{2},\frac{n}{2}) $$

Further, as $f_B(\cdot,\frac{n}{2},\frac{n}{2})$ is positive on $[0,1]$, we conclude that for
any interval $[\underline{p}',\overline{p}']$ with $\overline{p}'-\underline{p}'=\overline{p}-\underline{p}$ we have 
$\mathbb{P}\left(p \in \left[\underline{p}',\overline{p}'\right]\right)\leq \mathbb{P}\left(p \in \left[\underline{p},\overline{p}\right]\right)$.
As all steps in the chain above are equalities, $\left[\underline{p},\overline{p}\right]$ is indeed the smallest interval for which we can 
guarantee $\mathbb{P}\left(p \in \left[\underline{p},\overline{p}\right]\right) \geq 1-\delta_T$.

Next, we consider arbitrary $1\leq k\leq n-1$. In this case we construct the following interval which contains $p$ with probability $1-\delta_T$ by definition:

\[
    \left[\underline{p}_k,\overline{p}_k\right] = [I^{-1}_{\nicefrac{\delta_T}{2}}(k,n-k),I^{-1}_{1-\nicefrac{\delta_T}{2}}(k,n-k)]
\]

Note that for $k=\nicefrac{n}{2}$ the intervals $\left[\underline{p},\overline{p}\right]$ and $\left[\underline{p}_k,\overline{p}_k\right]$ coincide.

We now show that the size of the interval $\overline{p}_k-\underline{p}_k$ is maximal if $k=\nicefrac{n}{2}$. We do this by computing the derivative with respect to k.
Substituting $a=k$ and $b=n-k$, using symmetry, applying the multi-variable chain rule, and renaming integration variables we obtain

\begin{align*}
    & \frac{\partial}{\partial k}(I^{-1}_{1-\nicefrac{\delta_T}{2}}(k,n-k)-I^{-1}_{\nicefrac{\delta_T}{2}}(k,n-k))  \\
    &= \frac{\partial}{\partial k}(I^{-1}_{1-\nicefrac{\delta_T}{2}}(a,b)-I^{-1}_{\nicefrac{\delta_T}{2}}(a,b)) \\
    &= \frac{\partial}{\partial k}(1-I^{-1}_{\nicefrac{\delta_T}{2}}(b,a)-I^{-1}_{\nicefrac{\delta_T}{2}}(a,b)) \\
    &= \frac{\partial}{\partial b}I^{-1}_{\nicefrac{\delta_T}{2}}(b,a)-
    \frac{\partial}{\partial a}I^{-1}_{\nicefrac{\delta_T}{2}}(b,a)-
    \frac{\partial}{\partial a}I^{-1}_{\nicefrac{\delta_T}{2}}(a,b)+
    \frac{\partial}{\partial b}I^{-1}_{\nicefrac{\delta_T}{2}}(a,b) \\
    &= \frac{\partial}{\partial c}I^{-1}_{\nicefrac{\delta_T}{2}}(c,a)-
    \frac{\partial}{\partial c}I^{-1}_{\nicefrac{\delta_T}{2}}(b,c)-
    \frac{\partial}{\partial c}I^{-1}_{\nicefrac{\delta_T}{2}}(c,b)+
    \frac{\partial}{\partial c}I^{-1}_{\nicefrac{\delta_T}{2}}(a,c) \\
    &= \frac{\partial}{\partial c}\left(I^{-1}_{\nicefrac{\delta_T}{2}}(c,a)-I^{-1}_{\nicefrac{\delta_T}{2}}(c,b)\right)
    -\frac{\partial}{\partial c}\left(I^{-1}_{\nicefrac{\delta_T}{2}}(b,c)-I^{-1}_{\nicefrac{\delta_T}{2}}(a,c)\right) \\
    &= \frac{\partial}{\partial c}\left(I^{-1}_{\nicefrac{\delta_T}{2}}(c,b)-I^{-1}_{\nicefrac{\delta_T}{2}}(c,a)\right)
    +\frac{\partial}{\partial c}\left(I^{-1}_{1-\nicefrac{\delta_T}{2}}(c,b)-I^{-1}_{1-\nicefrac{\delta_T}{2}}(c,a)\right) \\
\end{align*}

Clearly, for $a=b$ the expression equals 0, i.e., for $k=\nicefrac{n}{2}$ the interval size reaches an extreme point.
As the function $a\mapsto I^{-^1}_p(a,b)$ for any $p\in (0,1)$ and $b\geq 1$ is concave on $[1,\infty)$ \cite{Askitis2021},
both $ \frac{\partial}{\partial c}(I^{-1}_{1-\nicefrac{\delta_T}{2}}(b,c)-I^{-1}_{1-\nicefrac{\delta_T}{2}}(a,c))$ and 
$\frac{\partial}{\partial c}(I^{-1}_{\nicefrac{\delta_T}{2}}(b,c)-I^{-1}_{\nicefrac{\delta_T}{2}}(a,c))$
are positive if and only if $b>a$, i.e., exactly when $\nicefrac{k}{2} < n$. Analogously, the expression is negative for $\nicefrac{k}{2} > n$.
Thus, $k=\nicefrac{n}{2}$ is the only extreme point in the interval $[0,1]$ and a maximum.

This means for every $k$ we have an interval $\left[\underline{p}_k,\overline{p}_k\right]$ that contains $p$ with probability at least $1-\delta_T$ and has size bounded by

$$ \overline{p}_k-\underline{p}_k \leq 1-2I_{\nicefrac{\delta_T}{2}}^{-1}\left( \frac{n}{2}+1,\frac{n}{2}+1\right)$$

and for $k=\nicefrac{n}{2}$ no smaller interval exists.
\end{proof}

Note that in \Cref{line:2} we assumed a uniform prior in the binomial distribution.
However, we can easily generalize this result for other choices of beta-distributed priors.
Observe that the proof only relies on the parameters of the beta distribution but not how the parameters
are composed, i.e., to which extent the prior hyperparameters or the samples contributed.
Thus, we can generalize the result as follows: 

\begin{corollary}\label{corollary:betaprior}
    Let $k \sim \mathit{Bin(n,p)}$ be a random variable according to a binomial distribution.
    Assume a a beta-distributed prior $p\sim B(\alpha_1,\alpha_2)$
    Then the smallest interval 
    $[\underline{p},\overline{p}]$ for which
    \[
    \mathbb{P}\left(p \in \left[\underline{p},\overline{p}\right]\right) \geq 1-\delta_T
    \]
    holds, has size
    \[
    \overline{p}-\underline{p} \leq 1-2I_{\nicefrac{\delta_T}{2}}^{-1}\left( \frac{n+\alpha_1+\alpha_2}{2},\frac{n+\alpha_1+\alpha_2}{2}\right).
    \] 
\end{corollary}

Note that the interval size decreases monotonically as $n+\alpha_1+\alpha_2$ increases. As $\alpha_1,\alpha_2>0$, in case no information about the prior distribution of $p$ is present, we can still give a lower bound of the interval size, namely $1-2I_{\nicefrac{\delta_T}{2}}^{-1}\left( \frac{n}{2},\frac{n}{2}\right)$ by underapproximating $\alpha_1=\alpha_2=0$.

\begin{proof}[Proof Theorem~\ref{thm:beta-zeta-bound}]
\hspace{0cm}\\
Let $\Mts = (\Smain \cup \Saux, A \cup \{\tau\}, P^\ts, R^\ts, \gamma)$ be the 2-successor MDP obtained by transforming $M$ as in \Cref{subsec:mdp_transformation}.
We then define the set of bootstrapped state-action pairs in $\Mts$ as $\mathfrak{B}^\ts=\{(s,a)\mid s\in \Smain, \data^\ts(s,a) < \Nmin^\beta \}
\cup \{(s,\tau)\mid s\in \Saux \}$. This means the set of non-bootstrapped actions is of size at most $|S||A|$, i.e., $\lvert\{(s,a)\mid (s,a)\not\in\mathfrak{B}^{2s}\}\rvert\leq |S||A|$. Distributing the error tolerance $\delta$ uniformly over all state-action pairs and by applying \Cref{corollary:betaprior} we can ensure with high probability $1-\delta$ that $e(s,a)\leq 1-2I_{\nicefrac{\delta_T}{2}}^{-1}\left( \frac{n}{2}+1,\frac{n}{2}+1\right)$ for all $(s,a)\not\in\mathfrak{B}$.
Note that although we assumed uniform priors for all transitions in $M$, we do not necessarily have uniform priors for all transitions in $\Mts$. Precisely, by the marginal distributions of the Dirichlet distribution, all transitions $(s,a)$ in $\Mts$ have a prior of $B(1,m)$ where $m$ is the number of states reachable from $s$ by action $a$ through only auxiliary states. This is why we have to apply \Cref{corollary:betaprior} rather than \Cref{lemma:binomial_bound}. However, the bound from \Cref{lemma:binomial_bound} is still as tight as possible since there are transitions with $m=1$, namely all $\tau$-transitions from the last auxiliary state after the transformation and all transitions that were already binary before the transformation.

We then finish this proof in the same fashion as the proof of \Cref{thm:2smdp_spi} and \Cref{lemma:2smdp_spi}, but use the above mentioned bound for the $L_1$ error instead of the bound obtained in \cite{Weissman}.
\end{proof}

\end{document}